\documentclass[sigconf,table]{acmart}

\setcopyright{none}
\renewcommand\footnotetextcopyrightpermission[1]{}
\usepackage{multirow}
\usepackage{comment}

\usepackage{tikz}
\usepackage{pgf-pie}
\usetikzlibrary{shapes,arrows}
\tikzstyle{decision} = [diamond, draw, fill=blue!20, 
    text width=4.5em, text badly centered, node distance=3cm, inner sep=0pt]
\tikzstyle{block} = [rectangle, draw, fill=blue!20, 
    text width=5em, text centered, rounded corners, minimum height=4em]
\tikzstyle{line} = [draw, -latex']
\tikzstyle{cloud} = [draw, ellipse,fill=red!20, node distance=3cm,
    minimum height=2em]

\AtBeginDocument{%
  \providecommand\BibTeX{{%
    \normalfont B\kern-0.5em{\scshape i\kern-0.25em b}\kern-0.8em\TeX}}}

\begin{document}

\title{A Systematic Literature Review on Client Selection in Federated Learning}


\author{Carl Smestad}
\affiliation{%
  \institution{Norwegian University of Science and Technology}
  \city{Trondheim}
  \country{Norway}}
\email{carl.smestad@gmail.com}

\author{Jingyue Li}
\affiliation{%
  \institution{Norwegian University of Science and Technology}
  \city{Trondheim}
  \country{Norway}}
\email{jingyue.li@ntnu.no}

\renewcommand{\shortauthors}{Smestad and Li}

\begin{abstract}
    With the arising concerns of privacy within machine learning, federated learning (FL) was invented in 2017, in which the clients, such as mobile devices, train a model and send the update to the centralized server. Choosing clients randomly for FL can harm learning performance due to different reasons. Many studies have proposed approaches to address the challenges of client selection of FL. However, no systematic literature review (SLR) on this topic existed.
    This SLR investigates the state of the art of client selection in FL and answers the challenges, solutions, and metrics to evaluate the solutions. We systematically reviewed  47 primary studies. The main challenges found in client selection are heterogeneity, resource allocation, communication costs, and fairness. The client selection schemes aim to improve the original random selection algorithm by focusing on one or several of the aforementioned challenges.
    The most common metric used is testing accuracy versus communication rounds, as testing accuracy measures the successfulness of the learning and preferably in as few communication rounds as possible, as they are very expensive.
    Although several possible improvements can be made with the current state of client selection, the most beneficial ones are evaluating the impact of unsuccessful clients and gaining a more theoretical understanding of the impact of fairness in FL.
\end{abstract}

\ccsdesc[500]{Computing methodologies~Distributed artificial intelligence}
\ccsdesc[500]{Computing methodologies~Neural networks}
\ccsdesc[300]{Computing methodologies~Distributed algorithms}

\keywords{systematic literature review, software metric, federated learning, client selection, neural network}

\maketitle

\section{Introduction}
Machine learning (ML) has increased in popularity in recent years amongst businesses and research. Also, cellphones and tablets are the primary computing devices for many people \cite{ anderson_technology_2015}. These devices are equipped with powerful sensors such as cameras, microphones, and GPS, resulting in a vast amount of private data. With this increased concern regarding personal- and data privacy, a new paradigm for machine learning arose named decentralized learning, with the most prominent technique being federated learning (FL).

FL was introduced in 2017 by \citet{mcmahan_communication-efficient_2017}, which is a decentralized ML paradigm that leaves the training data distributed on mobile devices and learns a shared model by aggregating locally-computed updates \cite{ma_state---art_2022}. Instead of sending private data to a centralized server (CS), the clients compute or train a model on their device and send the update to the centralized server. The server randomly selects a fixed-size  subset of clients and provides them with an initial global model before they train and send the updates. As the client devices have different data, randomly selecting them might lead to several challenges. 
As the FL models are meant to be trained on smartphones and IoT devices, the expensive cost of communication must be
considered by reducing the number of communication rounds and reducing the size of
the transmitted messages \cite{li_federated_2020}. 
Another culprit of FL is that it is performed synchronously, which implies
that one round of training is finished when every edge device in the network has sent its
model. This results in an effect known as the straggler effect, where the network is as fast as
the slowest edge \cite{qu_context-aware_2022}. Assuming there is an algorithm that selects the fastest clients and there are no slow clients inducing the straggler effect, and the communication cost and communication rounds are at a minuscule level, it would be natural to assume that this algorithm would be ideal and that not much more could be done in order to improve it. \textbf{However, client selection is much more complicated than that.} 

To ensure good learning, one must also consider the data heterogeneity, resource allocation, and fairness between the clients. Likely, the selected clients do not have the same data distribution, which might lead to heavy biases within the learning.
Thus, choosing the "best" clients is an integral part of a well-functioning
federated learning network, but it is certainly not trivial. There are many issues to consider for
defining the best client selection algorithm. 

As pointed out by \citet{mcmahan_communication-efficient_2017}, only a fraction of clients are selected for efficiency, as their experiments show diminishing returns for adding more clients beyond a certain point.
If the algorithm chooses to include every client as opposed to only a subset, there might be a lot of included clients who do not add value to the training while increasing the cost of communication. Perhaps some of the clients will not even finish training, which will make the entire round of learning fail. Therefore, only a subset of clients must be included during client selection.

Hence, we are motivated to examine how studies have tried to improve client selection, as many problems can be addressed by better strategies than random selection. We focus on answering the following research questions (RQs):

\begin{itemize}
    \item \textbf{RQ1: What are the main challenges in client selection?}
    \item \textbf{RQ2: How are clients selected in federated learning?}
    \item \textbf{RQ3: Which metrics are important for measuring client selection?}
    \item \textbf{RQ4: What can be improved with the current client selection?}
\end{itemize}

To answer the research questions, a systematic literature review (SLR) was conducted by following the guidelines by \citet{kitchenham_guidelines_2007} and \citet{wohlin_guidelines_2014}. One iteration of backward and forward-snowballing was performed on a set of six papers, resulting in 47 primary studies to review after the quality assessment and study selection. The contributions of this SLR are as follows.

\begin{itemize}
    \item  It summarizes the main challenges in terms of client selection for FL. The main challenges are heterogeneity, resource allocation, communication costs, and fairness.
    \item It summarizes the important metrics for measuring client selection in regard to the main challenges. The most commonly used metrics are testing accuracy and communication rounds.
    \item It discusses possible future work within the field of client selection for FL.
\end{itemize}

The rest of the paper is organized as follows. The related work is presented in section \ref{sec:related-work}. The research methodology and implementation are presented in section \ref{sec:research-design-and-implementation}. Section \ref{sec:research-results} shows the results of this SLR, and section \ref{sec:discussion} discusses the results. Lastly, section \ref{sec:future-work} concludes the study and proposes future work.

\section{Related Work}
\label{sec:related-work}

There are several literature reviews and surveys related to FL. \citet{hou_systematic_2021} performed an SLR of blockchain-based FL and specialized in the architectures and applications. They identified four security issues in FL which motivate the use of blockchain. The study mentioned the Internet of Things (IoT), medicine, and the Internet of Vehicles (IoV) as promising fields for application but did not mention client selection.

\citet{pfitzner_federated_2021} conducted an SLR of FL in a medical context. They focused on the areas that were promising for digital health applications. \citet{antunes_federated_2022} did an SLR of FL for healthcare and focused on the architecture and remaining issues regarding applying FL to electronic health records (EHR). Both \citet{pfitzner_federated_2021}  and \citet{antunes_federated_2022} focused on the security perspective and did not summarize client selection issues.
\citet{lo_systematic_2021} performed an SLR of FL from a software engineering perspective. They focused on what FL is, the different applications, general challenges, and how they are addressed. The five most common challenges were communication efficiency, statistical heterogeneity, system heterogeneity, data security, and client device security. The study noticed that client selection is mostly server-based but did not discuss it further.
\citet{liu_systematic_2020} conducted an SLR of FL but from a model quality perspective. The study presents several algorithms types, such as neural networks, decision trees, etc., with corresponding client-side algorithms but does not consider client selection. 
\citet{shaheen_applications_2022} investigated the applications, challenges and research trends of FL. The study underwent 105 research studies and discovered that the most promising application is within the healthcare domain. They reported data imbalance, system heterogeneity, expensive communication, privacy concerns, statistical heterogeneity, and resource allocation as the main challenges of implementing FL. However, they did not relate any of these challenges to client selection.
\citet{witt_decentral_2022} wrote an SLR of FL from the incentivization methods perspective. This study also discusses blockchain as a possible improvement but does not mention client selection outside the scope of blockchain.
\citet{hosseinzadeh_federated_2022} did an SLR of FL with emphasis on IoT, focusing on the evaluation factors and the future and open challenges of FL-based IoT. The study mentions a possible client selection method but does not focus on the topic.
\citet{lo_architectural_2022} reviewed the different architectural patterns to design FL systems. The study reports 15 architectural patterns, where one of which is the client selector. The study provides a high-level overview of possible solutions, such as resource-based, data-based, and performance-based client selection, as well as some of the benefits and drawbacks of the pattern.
\citet{abreha_federated_2022} systematically surveyed FL in edge computing. 
The survey reports the main challenges as communication cost, reliability, privacy, and administrative policies. It also discusses client selection to a small degree by mentioning existing studies on the topic.
\citet{ali_incentive-driven_2021} conducted an SLR of incentive-driven FL and the associated security challenges. Some incentive mechanisms include auction theory and blockchain but do not touch on the topic of client selection and possibly how to incentivize clients.
\citet{ma_state---art_2022} reviewed the state-of-the-art in solving non-\\independant and identically distributed (Non-IID) data in FL and addressed future trends for the topic. When the datasets are not independent and identically distributed, it leads to less correlation and dependencies because samples of the datasets do not have the same probability distribution. Non-IID data is one of the largest challenges in FL, and the study discusses ways to improve it through, e.g., data enhancements and data selection.
One of these methods is client selection, but the survey does not go more into depth than linking to relevant papers.

\section{Research Design and Implementation}
\label{sec:research-design-and-implementation}

To summarize the state of the art of client selection of FL and to answer our research questions, we performed a systematic literature review based upon the guidelines  \cite{kitchenham_guidelines_2007} and \cite{wohlin_guidelines_2014}.  

\subsection{Search Strategy}
\begin{figure}[H]
    \centering
    \includegraphics[width=0.4\textwidth]{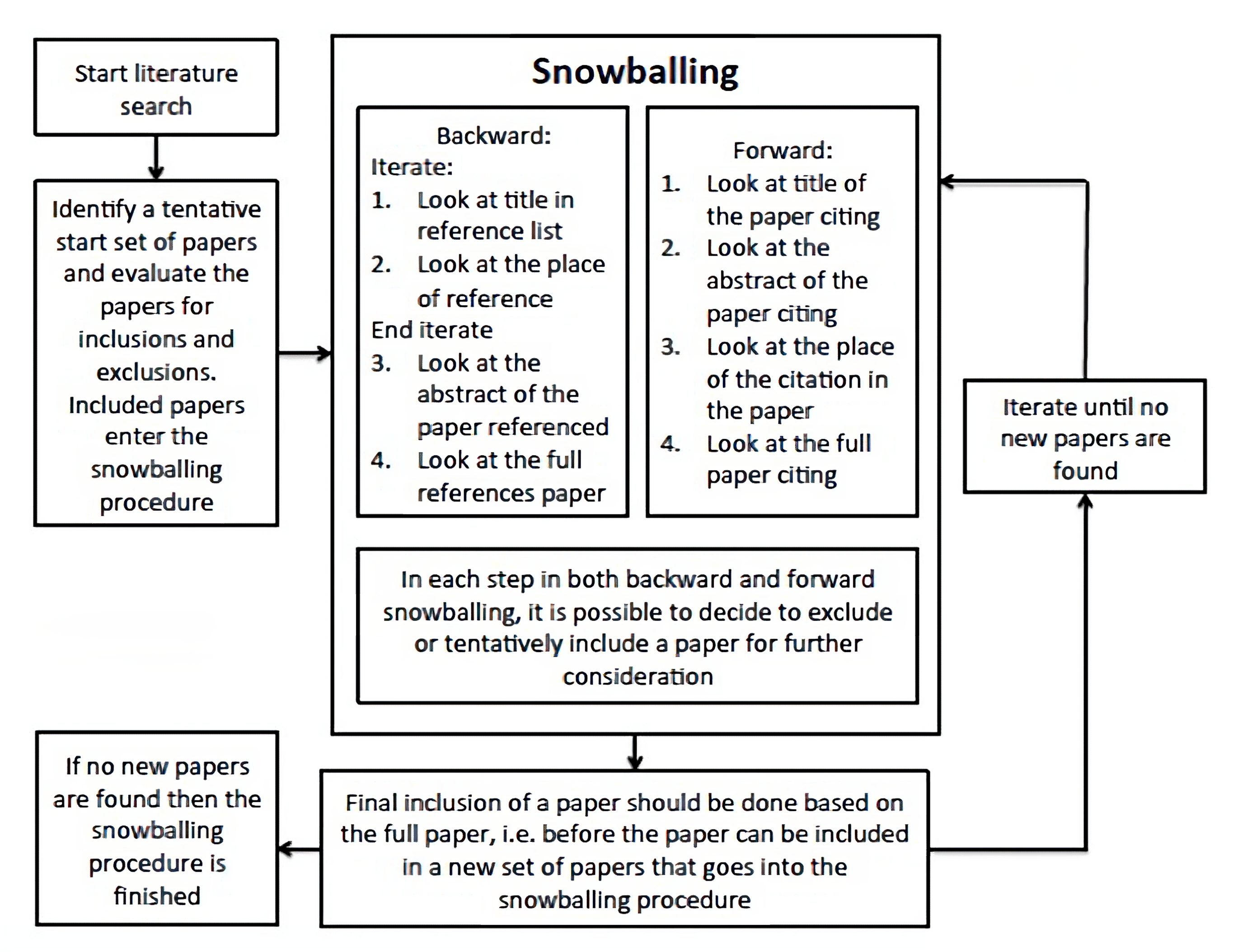}
    \caption{Illustration of snowballing in SLR \cite{wohlin_guidelines_2014}}
    \label{fig:snowballing}
\end{figure}

Generally, the SLR approach for generating a search strategy is to break down the research questions into searchable terms and generate a list of synonyms, abbreviations, and alternative spellings.
As there exist a vast amount of studies on the topic of FL, this process became unmanageable. Thus, the strategy used in this paper is based on the guidelines for snowballing in SLR by \citet{wohlin_guidelines_2014}, as shown in Figure
\ref{fig:snowballing}, which includes the following main steps:

\begin{itemize}
    \item Step 1: Generate a start set of studies (including only papers that will be a part of the final analysis)
    \item Step 2: Perform backward- and forward snowballing
    \item Step 3: Decide to include or exclude the study
    \item Step 4: Iterate until finding no new papers
\end{itemize}

To start the snowballing procedure, a starting set was needed. Google Scholar was used to generate this starting set by using relevant terms such as "Federated Learning" and "Client Selection in Federated Learning."
The results are listed  below.

\begin{itemize}
\item "Communication-Efficient Learning of Deep Networks from  Decentralized Data" \cite{mcmahan_communication-efficient_2017}  
\item "Client Selection for Federated Learning with Heterogeneous Resources in Mobile Edge" \cite{nishio_client_2019}   
\item "Client selection and bandwidth allocation in wireless federated learning networks: A long-term perspective" \cite{xu_client_2021} 
\item "Federated Learning in a Medical Context: A Systematic Literature Review" \cite{pfitzner_federated_2021} 
\item "A Systematic Literature Review of Blockchain-based Federated Learning: Architectures, Applications and Issues" \cite{hou_systematic_2021}  
\item "A state-of-the-art survey on solving non-IID data in Federated Learning" \cite{ma_state---art_2022}   
\end{itemize}

When starting to perform forward and backward snowballing on the starting set, it was apparent that there were too many papers to add as \cite{mcmahan_communication-efficient_2017} is the first paper on federated learning and is cited by almost every relevant paper in the field.
The paper provided a definition name for devices in FL, namely "clients." By investigating several studies, it was clear that, despite FL being a young field within machine learning, a consensus existed on using the term "Client Selection" for choosing the appropriate devices. Thus, a substring search on Google Scholar with the "cited by" feature was conducted with that substring to choose the most relevant studies.

\subsection{Study Selection and Quality Assessment}
We defined including and exclusion criteria to identify primary studies. For a paper to be included, it has to fulfil all the following inclusion  criteria: 
\begin{itemize}
\item Written in English
\item Published after 2017 because 2017 is the origin of federated learning appeared in \cite{mcmahan_communication-efficient_2017}
\item Discusses client selection in Federated Learning
\item Peer-reviewed 
\end{itemize}

According to \cite{higgins_cochrane_2019}, quality can be seen as to which extent the study minimizes bias and maximizes internal and external validity.
Table \ref{table:quality-assessment} shows the different quality assessment criteria for empirical and non-empirical sources \cite{mohanani_cognitive_2020}. For each selected paper, we assessed its quality according to the quality assessment criteria and awarded one point for \textit{yes} and zero points for \textit{no}. We awarded half a point if it was uncertain whether or not the study fulfilled the criterion. 
Then an average was generated for each paper. A paper has to have an average of 0.5 or more to be accepted as the primary study. 

By applying the selection and quality assessment criteria, a total of 47 papers were chosen as primary studies for data extraction and synthesis. 

\begin{table*}[h!]
\caption{Quality assessment criteria based on \cite{mohanani_cognitive_2020}}
\begin{tabular}{lccl}
\textbf{Quality Criteria}                              & \multicolumn{1}{l}{\textbf{Empirical}} & \multicolumn{1}{l}{\textbf{Non-empirical}} &  \\ \cline{1-3}
Was the motivation for the study provided?               & X                                      & X                                          &  \\
Is the relevance to the industry discussed?            & X                                      & X                                          &  \\
Are the most important sources linked to or discussed? & X                                      & X                                          &  \\
Is the aim (e.g., objectives, research goal) reported?  & X                                      & X                                          &  \\
Was the research method or design described?           & X                                      &                                            &  \\
Were any threats to validity clearly stated?     & X                                      &                                            &  \\
                                                       & \multicolumn{1}{l}{}                   & \multicolumn{1}{l}{}                       &  \\
                                                       & \multicolumn{1}{l}{}                   & \multicolumn{1}{l}{}                       & 
\end{tabular}
\label{table:quality-assessment}
\end{table*}

\subsection{Data Synthesis}
\label{sec:data-extraction}

Data synthesis involves collating and summarizing the results of the primary studies \cite{kitchenham_guidelines_2007}. SLRs within IT and software engineering are generally qualitative in nature.
Based on the overview of data synthesis provided by \cite{van_den_berg_overview_2013}, we synthesize the data in a spreadsheet, where the common themes, patterns, and finding between the extracted information can be viewed. 

For each RQ, relevant data were extracted and put into their respective columns according to the research question.
Lastly, a list was manually generated based on the challenges and themes that were created for answering the research questions. The data synthesis process was recorded and available at \footnote{\url{https://docs.google.com/spreadsheets/d/1jIGpbkOcXazFRcR_Rds0mTshX0NDCIXU9Wgw3SECAiw/edit?usp=sharing}}.

\section{Research Results}
\label{sec:research-results}

This section presents the results of each research question.


\subsection{RQ1: What are the main challenges in client selection?}
\label{sec:result-rq1}

\begin{figure}[ht]
    \centering
    \begin{tikzpicture}[scale=0.7]
        \pie[sum=auto]{
            23/Heterogeneity,
            13/Resource Allocation,
            8/Communication Costs,
            3/Fairness}
    \end{tikzpicture}
    \caption{Distribution of challenges reported from the primary studies}
    \label{fig:pie-chart-challenges}
\end{figure}
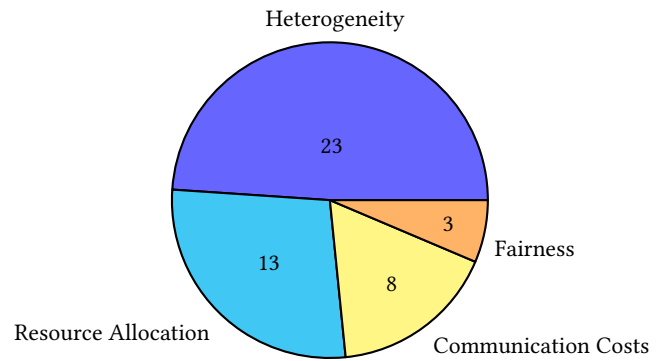

Results show that 23 studies tried to improve upon heterogeneity, 13 studies revolved around resource allocation, eight studies focused on communication costs, and three studies had fairness as the main challenge.
The distribution of the challenges can be seen in Figure \ref{fig:pie-chart-challenges}. Several studies report more than one challenge, but it has been assigned to the challenge it focuses mostly on.

\subsection{Heterogeneity}
\label{sec:result-rq1-heterogenity}
In FL, the training is executed on the client's local devices. This will result in differences between the clients as they will have different datasets and availability.
This is the most common challenge found in FL, and \citet{mcmahan_communication-efficient_2017} reported heterogeneity as the main challenge. Almost half of the primary studies tried to improve it through different measures.

\citet{ma_state---art_2022} conducted a state-of-the-art survey on solving non-IID data in FL and concluded that data heterogeneity could be divided into the following categories: feature distribution skew, label distribution skew, same label (different features), same feature (different labels) and quantity skew. \citet{cho_towards_2022} report that heterogeneity also might arise due to partial client participation, as only a small fraction of client nodes participate in each round of training. 
If the client selection algorithm selects an improper subset of clients with poor-quality data, this will result in an inefficient trained model \cite{saha_data-centric_2022}.
\citet{ma_client_2021} reported label distribution skew as one of the most significant parameters which lead to performance degradation, while \citet{rai_client_2022} reported skewed data as one of the most critical factors. \citet{zhang_client_2021} reported that heterogeneity / Non-IID might bring the biases of some clients into the model training and cause accuracy degradation. This claim is supported by \citet{zhang_dubhe_2021}, who claim an urgent need for client selection strategies that promise data unbiasedness in FL. \citet{li_pyramidfl_2022} analyzed the limitations of the state-of-the-art client selection in regard to heterogeneity and concluded that due to under-exploited statistical- and system efficiency, not all the model updates would contribute to the model training equally. As various clients have diverse data sizes and importance, uploading unimportant updates significantly degrades the system's efficiency. According to \citet{li_data_2022}, a significant problem with utilizing FL with IoT is that the local data of sensors are constantly changing. This will have a similar effect as device failures and might lead to skewed distributed data, which leads to model degradation. There might also exist label noise on some clients, which exists naturally. This will lead to unnecessary information being exchanged \cite{yang_client_2022}.

To summarize, the key findings for the challenge of heterogeneity are as follows.
\begin{itemize}
    \item 48.93\% of the studies reported heterogeneity as the main challenge for FL.
    \item It might result in an inefficient trained model, performance- and accuracy-degradation.
    \item Heterogeneity might increase biases and unnecessary exchange of information. 
\end{itemize}

\subsection{Resource Allocation}
\label{sec:result-rq1-resource-allocation}
Resource allocation was the second most common problem in the primary studies.
This is due to several reasons, but the main one is due to the fact that the training process becomes inefficient when some clients have limited computational resources \cite{nishio_client_2019}.
\citet{xu_client_2021} state that a considerable challenge in resource allocation is that learning rounds are temporally interdependent and have varying significance toward the final learning outcome.
According to \citet{yu_jointly_2022}, it is unnecessary to select more clients than needed, and it is beneficial to have fewer clients. Still, the challenge consists of the trade-off between the number of clients, energy consumption, and resource allocation. Furthermore, within hierarchical federated learning (HFL), unique challenges exist, such as clients sometimes being inaccessible to the edge servers. 
Due to differences in resources and hardware specifications, the "straggler effect" is bound to happen \cite{qu_context-aware_2022}. \citet{zhang_adaptive_2021} stated that clients are constrained by personal energy and computation that may reduce the efficiency of ML training tasks. This is because the training and transmission of large models are very energy-consuming and might be difficult on low-energy edge devices. During training, there might be changes in client resources due to volatility of client population, client data, and training status \cite{shi_vfedcs_2022}. The topic of energy consumption within FL is important, as training and transmission of large models are energy-consuming, while edge devices generally have little energy. \citet{zeng_energy-efficient_2020} propose a client selection policy of giving the lowest priority to clients with poor communication capacity and a bad channel.

To summarize, the key findings for the challenge of resource allocation are as follows.
\begin{itemize}
    \item 27.65\% of the studies reported resource allocation as the main challenge of FL.
    \item The training process becomes inefficient when some clients have limited computational resources.
    \item Training and transmission of large models are very energy-consuming and difficult for low-energy devices.
\end{itemize}

\subsection{Communication costs}
\label{sec:result-rq1-communication-costs}

The third most common problem was the communication costs in FL. The communication cost is essential as every time the global model updates, it needs to receive the local aggregation of all the selected clients. According to \citet{tan_reputation-aware_2022}, the communication power required to reach convergence makes up a large portion of the cost.
One of the challenges is that a client with low computing power might not return the local model update on time, leading to a long convergence time \cite{ko_joint_2021}.
Studies \cite{hosseinzadeh_communication-loss_2022, li_uncertainty_2022} state that the trade-off between communication costs and accuracy is a challenge. \citet{asad_thf_2022} state that another challenge is the long distance between the different clients and the global server, which results in increased bandwidth usage. 

By default, FL is done synchronously. This implies that a round of communication / global model updates is only executed once every client has uploaded their model. This leads to an effect known as the straggler effect, where the system is only as fast as the slowest link \cite{zhu_client_2022}. This issue is also addressed by \citet{qu_context-aware_2022}.

Another fundamental challenge with communication costs is the energy usage of clients in FL. As vast amounts of data are generated from mobile and edge devices, these devices are energy-restricted. It is imperative to improve the energy efficiency of the systems \cite{zeng_energy-efficient_2020}.

According to \citet{deng_auction_2022}, clients' hardware conditions and data resources can vary significantly, which might lead to negative performance.

To summarize, the key findings for the challenge of communication costs are as follows.
\begin{itemize}
    \item 17.02\% of the studies reported communication costs as the main challenge of FL.
    \item Clients with low power or slow will lead to long convergence time. As FL is done synchronously, this implies that the learning is as fast as the slowest client.
    \item The possibly long distance between clients and servers will result in increased bandwidth usage.
\end{itemize}

\subsection{Fairness}
\label{sec:result-rq1-fairness}

The last common problem encountered was fairness. Only three studies reported it as the main challenge which they tried to solve. However, fairness is a researched topic within several similar fields, such as Resource Allocation (RA) and ML. In the context of resource allocation, the problem is defined as allocating a scarce shared resource among many users. For machine learning, it is typically defined as the protection of some specific attribute(s) by, e.g., preprocessing the data to remove information about the protected attribute \cite{feldman_computational_2015}.

In the context of FL, if the client selection algorithm always selects the fastest devices, it might boost the training process. However,  as stated by \citet{huang_efficiency-boosting_2021} " \textit{But clients with low priority are simply being deprived of chances to participate at the same time, which we refer to it as an unfair selection.}"
It might result in undesirable effects, such as omitting some portions of data. Also, if there are less data involved, data diversity will not be guaranteed and might hurt the performance of model training.
\citet{jee_cho_bandit-based_2020} state that by focusing on improving fairness, the uniformity of performance across clients will be improved as well. \citet{li_fair_2020} define fairness in FL as follows:

\begin{quote}
    \textbf{Definition 1} \textit{(Fairness of performance distribution)}. For trained models \textit{w} and \textit{$\tilde{w}$},
    $w$ provides a more fair solution to the federated learning objective \eqref{eq:Goal} than model \textit{$\tilde{w}$}, if the
    performance of model \textit{w} on the \textit{m} devices, $\{a1, \dots, a_m\}$, is more uniform than the performance of
    model \textit{$\tilde{w}$} on the \textit{m} devices.
\end{quote}

\begin{quote}
Note: Decoupling is the main benefit of FL. The FL algorithms may involve hundreds to millions of remote devices learning locally by minimizing the objective function $f(w)$ (1) \cite{li_fair_2020}:

\begin{equation}\label{eq:Goal}
    \textrm{min}_{w} f(w) = \sum_{k=1}^{m} p_{k} F_{k}(w)
\end{equation}

where \textit{m} is the total number of devices, $p_k \geq 0$, $\sum_k p_k = 1$, and the local objective $F_k$'s can be defined by empirical risks over local data.
\end{quote}

Through this definition, it becomes apparent that learned models which might be biased towards devices with large numbers of data points or commonly occurring devices are unfair.

According to \citet{ma_state---art_2022}, differences in data distribution and uncertainty in data quality are challenging in FL and data selection might exacerbate the unfairness of FL.
There are several different methods for prioritizing clients. If one selects all the "fast" devices, it might result in faster training but will deprive slower clients of the chance to participate. If the selection is one-sided, it will bring negative side effects, such as neutralizing some portions of data \cite{huang_efficiency-boosting_2021}.  In addition, clients may not provide honest results through various attacks, such as Byzantine attacks, which minimizes the effect of actual results of honest clients and reduces fairness \cite{wan_shielding_2022}.

To summarize, the key findings for the challenge of fairness are as follows.
\begin{itemize}
    \item 6.38\% of the studies reported fairness as the main challenge of FL.
    \item Selecting only the fastest clients might result in an unfair selection, as slower clients are deprived of the chance to participate.
    \item  An unfair selection might lead to heavy biases as some portions of the data are neutralized.
\end{itemize}


\subsection{RQ2: How are clients selected in federated learning}
\label{sec:result-rq2}
The different solutions are presented in this subsection and divided into their respective challenges. A summary of the findings is shown in Table \ref{table:challenges-solutions}.

\subsubsection{Heterogeneity} 

The most common approach to address this issue is to try to select a subset of clients who together give a more homogenous dataset \cite{cho_towards_2022}. \citet{ma_state---art_2022} performed a state-of-the-art survey on solving non-IID data in FL and mentioned \cite{wang_novel_2020} as a possible solution through client selection. 
They proposed selecting clients with small data heterogeneity based on Thompson sampling. \citet{abdulrahman_fedmccs_2021} suggested a similar algorithm of selecting a subset of clients who together form a homogeneous subset. \citet{zhang_client_2021} proposed to measure the degrees of non-IID data present in each client and then select the clients with the lowest degrees. \citet{li_pyramidfl_2022} and \citet{saha_data-centric_2022} had similar ideas but suggested a more holistic approach by also including the system heterogeneity (e.g., resources) as well. \citet{lin_contribution-based_2022} propose to dynamically update the selection weights according to the impact of the client's data.

Clustered Federated Learning (CFL) was introduced as an efficient scheme to balance out the non-IID data, and \citet{albaseer_client_2021} suggest leveraging the devices' heterogeneity to schedule them based on round latency and bandwidth to select clients. According to \citet{li_data_2022}, this type of approach works well within IoT due to the advantage of naturally clustered factory devices.
\citet{lee_data_2022} also find clusters of clients who together have near IID data through being distribution-aware.
In order to address the issue of label distribution skew, \citet{ma_client_2021} suggested a method where you check the similarity between the aggregated data distribution of the selected clients and compare it to the global data distribution. \citet{rai_client_2022} suggest giving each client an irrelevance score which improves the data distribution skewness. \citet{cao_c2s_2022} have an interesting approach to clustering clients by grouping them according to classes of data and then randomly selecting one client within every group. Another promising approach suggested by \citet{balakrishnan_diverse_2022} is to introduce diversity into client selection by measuring how a subset of clients can represent the whole when aggregated on the server for each communication round.

Generally, the studies try to keep an unbiased client selection in order to promote fairness. However, \citet{cho_towards_2022} report that biasing the client selection towards choosing clients with higher local losses resulted in an improvement in the partial client participation problem. \citet{abdulrahman_fedmccs_2021} suggested another approach to the same problem but suggested a multicriteria-based approach to predict if they were capable of performing the FL task.

Other studies, such as \cite{zhang_dubhe_2021}, suggest strengthening client selection with cryptographic methods such as homomorphic encryption (HE).
\citet{pang_incentive_2022} bring forward the idea of selecting clients at different global iterations to guarantee the completion of the FL job. Lastly, \citet{guo_wcl_2021} take into account both model weight divergence and local model training loss for selecting clients.

\subsubsection{Resource Allocation}
In order to improve the effect of some clients having limited resources, \citet{nishio_client_2019} suggest an algorithm that manages clients based on their resource conditions. Thus, allowing as many client updates as possible. \citet{xu_client_2021} create an algorithm that utilizes bandwidth allocation under long-term client energy constraints by using available wireless channel information in order to improve resource allocation. To deal with the resource allocation problem, \citet{yu_jointly_2022} suggest maximizing the number of clients while minimizing the energy consumption by the clients by allocating a set amount of resources in terms of CPU and transmission power.

Within HFL, \citet{qu_context-aware_2022} propose a client selection scheme with a network operator that learns the number of successful participating clients while dealing with a limited resource budget. Similarly,  \cite{deng_auction_2022, kuang_client_2021, liu_uplink_2022} suggested evaluating the learning quality of clients on a limited resource budget and then selecting the best clients. \citet{shi_vfedcs_2022} suggest that clients should be selected by considering and quantifying factors such as the relative impact of clients' data and resource differences and then selecting the clients with the most significant score.

Another method to deal with resource allocation is to focus on minimizing energy consumption and training delays in order to encourage more clients to participate in model updating. This may be done through reinforcement learning that learns to select the best subset of clients \cite{zhang_adaptive_2021}. \citet{du_-device_2022} propose an algorithm that utilizes fuzzy logic by considering the number of local data, computing capability, and network resources of each client.

\subsubsection{Communication Costs}
As communication cost is a vital challenge in FL, many attempts have been executed in order to improve it.
\citet{ko_joint_2021} developed a joint client selection algorithm that selects appropriate devices and allocates suitable amounts of resources to reduce convergence time due to high communication costs.
\citet{hosseinzadeh_communication-loss_2022} suggested a distributed client selection algorithm where the client devices participate in aggregation, resulting in lower communication costs while maintaining the low loss. \citet{li_uncertainty_2022} had a similar approach where they selected a subset of clients to participate in each round of training, and the remaining clients did not have to do any training, resulting in both lower computing and communication resources.

Another proposed solution is proposed by \citet{asad_thf_2022}, where there is a 3-way hierarchical framework to improve communication efficiency. It creates a cluster head that is responsible for communication with the global server, and local devices communicate with the cluster head. This will lead to model downloading and uploading requiring less bandwidth due to the short distances from the source to the destination. To tackle the energy consumption challenge, \citet{zeng_energy-efficient_2020} suggested only selecting the clients who provide significant information with each round. This would enable them to select fewer clients and end up with lower total energy consumption. In order to omit the "straggler effect" introduced through synchronous FL, \citet{zhu_client_2022} suggest an asynchronous approach where the server did not have to wait for all clients to be finished with their training. \citet{tan_reputation-aware_2022} proposed to utilize stochastic integer programming that selects clients in a reputation-aware manner.

\subsubsection{Fairness}
\citet{huang_efficiency-boosting_2021} promote a fairness-guaranteed client selection algorithm. They conclude that the final accuracy may increase by focusing on fairness but might sacrifice training efficiency. Whereas \citet{jee_cho_bandit-based_2020} suggest improving fairness through biased client selection by selecting the ones with higher local loss.
\citet{wan_shielding_2022} propose to select the most honest and useful clients by utilizing a multi-armed bandit approach, resulting in dishonest clients being filtered out.

\begin{table*}[h!]
\caption{Solutions compared to challenges}
\begin{tabular}{c|l}
\textbf{Challenge}  & \multicolumn{1}{c}{\textbf{Solution(s)}}                                                    \\ \hline
Heterogeneity        & \begin{tabular}[c]{@{}l@{}}- Select subset of client to make up homogeneous dataset\\ - Measure degrees of non-IID data and select lowest values\\ - Balance out non-IID data through clustered FL\\ - Give clients an irrelevance score and base selection of that\\ - Select a subset of clients who represent the entire set\\ - Utilize cryptography and weight divergence\end{tabular} \\ \hline
Resource Allocation & \begin{tabular}[c]{@{}l@{}}- Base selection on resource conditions\\ - Maximize the amount of clients by minimizing energy consumption\\ - Encourage clients to participate in model updating\\ - Utilize fuzzy logic by considering several resource factors\end{tabular}                                                                                                                                      \\ \hline
Communication Costs & \begin{tabular}[c]{@{}l@{}}- Joint client selection algorithm to reduce convergence time\\ - Distributed client selection where the clients decide to participate\\ - Only active clients should perform training\\ - 3-way hierarchical framework to improve efficiency\\ - Select the client with the most significant information each round\\ - Asynchronous FL\end{tabular}                \\ \hline
Fairness & \begin{tabular}[c]{@{}l@{}}- Fairness-guaranteed client selection algorithm\\ - Improve fairness through biased client selection\\ - Select honest clients\end{tabular}                                                                                                                                                                                                                                                                  
\end{tabular}
\label{table:challenges-solutions}
\end{table*}


\subsection{RQ3: Which metrics are important for measuring client selection?}
\label{sec:result-rq3}

The relevant metrics regarding client selection entirely depend on the problem the study is trying to improve upon. The different key metrics for each of the main challenges in client selection are summarized in Table \ref{table:metrics-challenges}.

\subsubsection{Heterogeneity}

The most common metric used is measuring the test accuracy against the number of communication rounds. Out of the 20 studies which reported heterogeneity as the biggest challenge,  14 used this metric to measure the success of their client selection. This metric was also utilized by the original FL paper \cite{mcmahan_communication-efficient_2017} and is directly comparable to the standard within regular machine learning, where "Test Accuracy vs. Epoch" is very commonly seen.
The main difference stems from FL having many clients send their model updates to a global server and then aggregate them. In that regard, a communication round corresponds to one epoch of the global server.

Studies \cite{cao_c2s_2022, li_data_2022} included a similar metric: the number of communication rounds up to a given threshold accuracy. This approach's main benefit is that it focuses more on minimizing the number of communication rounds, which are very costly in FL. Lastly, \citet{abdulrahman_fedmccs_2021} looked into how many selected clients are able to finish training without dropping out.

\subsubsection{Resource Allocation}
For the challenge of resource allocation, the most common metric seen is also "Testing Accuracy vs. Communication Rounds." This is as expected, as it directly measures how well the FL-algorithm performs.

Some studies supplement it with other metrics such as energy, delay, and client consumption \cite{zhang_adaptive_2021}. For mobile edge computing (MEC) systems, the energy is the basis of the client training model, and delay determines the iteration speed and convergence speed of the global model.

\subsubsection{Communication costs}
As already stated in section \ref{sec:result-rq1-communication-costs}, the cost of communication between clients and the global server is one of the most expensive parts of FL. Thus, utilizing the right metrics to validate the reduced cost is vital.

The typical "Testing Accuracy vs. Communication Rounds" is commonly seen in the studies, as higher testing accuracy in fewer communication rounds will lead to lower costs. Another beneficial metric is convergence time and latency, reported by \citet{ko_joint_2021}, as reducing the time spent in communication will lead to lower costs. 
Furthermore, \citet{tan_reputation-aware_2022} introduced the cost of hiring clients as an essential metric, as it was simply overlooked in existing studies and contributed a large part of the overall costs.

\subsubsection{Fairness}
Other than the already discussed "testing accuracy vs. communication rounds" metric, \citet{huang_efficiency-boosting_2021} utilized different metrics for measuring improved fairness.
For instance, they included metrics such as the availability of the client and mathematically measured the long-term fairness through constraints.

\begin{table*}[h!]
\caption{Metrics compared to challenges}
\begin{tabular}{c|l}
\textbf{Challenge}  & \multicolumn{1}{c}{\textbf{Metric(s)}}                                                                                                                                                      \\ \hline
Heterogeneity        & \begin{tabular}[c]{@{}l@{}}- Testing accuracy vs communication rounds\\ - Communication rounds until threshold accuracy\\ - Number of selected client able to finish training\end{tabular} \\ \hline
Resource Allocation & \begin{tabular}[c]{@{}l@{}}- Testing accuracy vs communication rounds\\ - Energy, delay, and client consumption\end{tabular}                                                                 \\ \hline
Communication Costs & \begin{tabular}[c]{@{}l@{}}- Testing accuracy vs communication rounds\\ - Convergence time vs latency\\ - Cost of hiring clients\end{tabular}                                               \\ \hline
Fairness            & \begin{tabular}[c]{@{}l@{}}- Testing accuracy vs communication rounds\\ - Availability of clients\\ - Long-term fairness constraints\end{tabular}                                          
\end{tabular}
    \label{table:metrics-challenges}
\end{table*}

\subsection{RQ4: What can be improved with the current client selection?}
\label{sec:result-rq4}

There are a lot of improvements that can be made with the current client selection. As decentralized learning is still pretty young, there is room for improvement within all discussed challenges in this SLR. 

\subsubsection{Heterogeneity}
For hierarchical federated learning (HFL), \citet{albaseer_client_2021} suggested looking into finding the optimal thresholds for splitting clusters of clients, which certainly would improve the communication efficiency of the learning network.

\subsubsection{Resource Allocation}
The primary studies reported several measures for possible future work which seem exciting and beneficial for the current client selection schemes.
\citet{xu_client_2021} stated that selecting clients as late as possible improves the efficiency of the client selection, but there is a lack of theoretical and practical research on the topic.

The most commonly suggested improvement was reported by three of the studies \cite{shi_vfedcs_2022, lin_contribution-based_2022, rai_client_2022}. They suggested looking into the effect of unsuccessful clients (or free-riders) and how to quantify the impact. These types of clients bring a lot of overhead costs into the learning network, and exploring the effects and solutions to those would undoubtedly improve the current client selection. 
None of the primary studies focused mainly on the effect of unsuccessful clients. However, studies \cite{huang_stochastic_2022, shi_vfedcs_2022} focused on optimizing client selection for volatile FL. This volatility stems from the dynamic of clients' data and the unreliable nature of clients (e.g., unintentional shutdown and network instability). Therefore, some work has been done on the topic, but there is certainly a gap that may be improved more. Those studies focus much more on the client's ability to enter and leave training rather than the effect of unsuccessful clients.

\subsubsection{Communication Costs}
\citet{ko_joint_2021} discussed the possibility of creating an incentive mechanism to encourage more computing power to FL. So far, there are no incentives for the client devices to allocate more resources to learning than necessary. Thus, giving them some sort of incentive mechanism would increase computational power and improve the problem of resource allocation.
Perhaps it would make it easier to create a more homogenous resource distribution amongst the clients.

\subsubsection{Fairness}
Even though only two studies reported fairness as the main challenge, studies, such as \cite{jee_cho_bandit-based_2020, zhu_client_2022, cho_towards_2022}, mention it as a possibly important factor that could promise a higher accuracy. Others mentioned it as a possible future direction for their work. For instance, \citet{shi_vfedcs_2022} reported that fairness might play an essential role in FL training and that studying it in a volatile context would be beneficial.
As already discussed in section \ref{sec:result-rq2}, studies already focus on fairness in client selection, but there is still a knowledge gap within the topic. \citet{huang_efficiency-boosting_2021} looked into the trade-off between fairness and training accuracy but concluded that they could not quantify the relationship and that looking further into analyzing the fairness factor for FL would be worthy of investigation.

\section{Discussion}
\label{sec:discussion}

This section discusses how the SLR compares to the related work as well as the limitations of the study.

\subsection{Comparison to Related Work}

To our knowledge, there currently does not exist any SLR focusing solely on client selection. The previous work has focused either on general FL challenges or the application of FL. Therefore, the main benefit of this  SLR is its focus on FL from the perspective of client selection.
However, there are a lot of similarities between the related work and this review, as they all encompass the challenges within FL. The value of this review to the industry is as a reference for the different client selection techniques and how they impact overall learning. There is also value in viewing possible future directions for client selection when looking into what can be improved.

This review has found a couple of areas that researchers may look more into from the perspective of client selection. Firstly, there are a vast amount of different client selection schemes proposed for FL, which all claim to outperform the state-of-the-art of random selection. It would be beneficial to compare these selection schemes with possible application cases in order to form an improved state-of-the-art solution for client selection. 
Secondly, the topic of fairness is not thoroughly explored. Several studies mention fairness as an important factor, but there does not exist much research on the topic of exploring the trade-offs and benefits of focusing on it.

Although FL is a relatively new field within machine learning, it already shows promising prospects within several application domains, such as healthcare, natural language processing, smart cities and IoT.
For certain industries, it might be more trivial to implement a well-functioning system as the developers know the types of devices on which the algorithm will be implemented, but this is not the case for applications such as IoT and edge computing.
In those fields, the developers do not necessarily know much about the client devices which will perform the learning, thus making it much more difficult to tackle the several challenges reported by the related work and found in this SLR. 

Client selection is an integral part of a  well-functioning FL system, as it may be utilized to improve the challenges of heterogeneity, resource allocation, communication costs, and fairness.
Despite the previous- and related work conducted on the topic, there is no de facto standard for the client selection algorithm within any application of FL. 
Even within a subset of any challenge, such as the issue of clients dropping out during training, multiple possible solutions exist, such as asynchronous FL, partial aggregation of dropped-out clients, and resource-aware FL. Within each category, there exist many algorithms to tackle the challenge through client selection, which shows the importance of exploring the topic further and possibly finding the best approach.

For academia and industry, this SLR may assist in several ways. Firstly, it can be used as a reference guide for the most prominent existing challenges and their consequences for learning. Secondly, for each given challenge, the SLR presents several different possible existing solutions to tackle it. This is especially valuable to the industry when deciding to implement an FL system and deciding whether or not their ecosystem is well-suited for it. The SLR also provides guidelines to mitigate some of the challenges.

\subsection{Limitations}
Although the guidelines for systematic reviews by \cite{kitchenham_guidelines_2007} were followed, several points may have been improved. We might miss some primary studies in the study search stage because there were many studies on the topic of FL. For instance, performing the forward snowballing procedure on the original FL-paper \citet{mcmahan_communication-efficient_2017} resulted in around 7000 studies. Even though there is plenty of academic research on the topic, we did not look into any grey literature as a possible source. There certainly may exist many exciting discussions and ideas on FL which are not discussed in academic journals but in blogs and newspapers. We might exclude papers that are relevant to the study during the paper selection process. To mitigate this risk, the papers' inclusion and exclusion are cross-checked and agreed upon by both authors.

\section{Conclusions and Future Work}
\label{sec:future-work}
We performed an SLR and summarized the challenges, solutions, and metrics for evaluating the solutions and possible future work of client selection in FL. Information from 47 primary studies is analyzed and synthesized. This study is the only SLR, as far as the author is aware, focusing solely on client selection in FL. 

The SLR delights several possible future research challenges we want to focus on. The most beneficial ones regard the impact of unsuccessful clients or fairness. Improving one of those challenges could benefit FL, as the training efficiency  might increase, and the communication costs would be reduced. The communication cost is also one of the most significant problems in FL. Thus, improving it would be beneficial.

\bibliographystyle{ACM-Reference-Format}
\bibliography{main}


\begin{thebibliography}{61}


\ifx \showCODEN    \undefined \def \showCODEN     #1{\unskip}     \fi
\ifx \showDOI      \undefined \def \showDOI       #1{#1}\fi
\ifx \showISBNx    \undefined \def \showISBNx     #1{\unskip}     \fi
\ifx \showISBNxiii \undefined \def \showISBNxiii  #1{\unskip}     \fi
\ifx \showISSN     \undefined \def \showISSN      #1{\unskip}     \fi
\ifx \showLCCN     \undefined \def \showLCCN      #1{\unskip}     \fi
\ifx \shownote     \undefined \def \shownote      #1{#1}          \fi
\ifx \showarticletitle \undefined \def \showarticletitle #1{#1}   \fi
\ifx \showURL      \undefined \def \showURL       {\relax}        \fi
\providecommand\bibfield[2]{#2}
\providecommand\bibinfo[2]{#2}
\providecommand\natexlab[1]{#1}
\providecommand\showeprint[2][]{arXiv:#2}

\bibitem[Abdulrahman et~al\mbox{.}(2021)]%
        {abdulrahman_fedmccs_2021}
\bibfield{author}{\bibinfo{person}{Sawsan Abdulrahman}, \bibinfo{person}{Hanine
  Tout}, \bibinfo{person}{Azzam Mourad}, {and} \bibinfo{person}{Chamseddine
  Talhi}.} \bibinfo{year}{2021}\natexlab{}.
\newblock \showarticletitle{{FedMCCS}: {Multicriteria} {Client} {Selection}
  {Model} for {Optimal} {IoT} {Federated} {Learning}}.
\newblock \bibinfo{journal}{\emph{IEEE Internet of Things Journal}}
  \bibinfo{volume}{8}, \bibinfo{number}{6} (\bibinfo{date}{March}
  \bibinfo{year}{2021}), \bibinfo{pages}{4723--4735}.
\newblock
\showISSN{2327-4662}
\urldef\tempurl%
\url{https://doi.org/10.1109/JIOT.2020.3028742}
\showDOI{\tempurl}
\newblock
\shownote{Conference Name: IEEE Internet of Things Journal}.


\bibitem[Abreha et~al\mbox{.}(2022)]%
        {abreha_federated_2022}
\bibfield{author}{\bibinfo{person}{Haftay~Gebreslasie Abreha},
  \bibinfo{person}{Mohammad Hayajneh}, {and} \bibinfo{person}{Mohamed~Adel
  Serhani}.} \bibinfo{year}{2022}\natexlab{}.
\newblock \showarticletitle{Federated {Learning} in {Edge} {Computing}: {A}
  {Systematic} {Survey}}.
\newblock \bibinfo{journal}{\emph{Sensors}} \bibinfo{volume}{22},
  \bibinfo{number}{2} (\bibinfo{date}{Jan.} \bibinfo{year}{2022}),
  \bibinfo{pages}{450}.
\newblock
\showISSN{1424-8220}
\urldef\tempurl%
\url{https://doi.org/10.3390/s22020450}
\showDOI{\tempurl}
\newblock
\shownote{Number: 2 Publisher: Multidisciplinary Digital Publishing Institute}.


\bibitem[Albaseer et~al\mbox{.}(2021)]%
        {albaseer_client_2021}
\bibfield{author}{\bibinfo{person}{Abdullatif Albaseer},
  \bibinfo{person}{Mohamed Abdallah}, \bibinfo{person}{Ala Al-Fuqaha}, {and}
  \bibinfo{person}{Aiman Erbad}.} \bibinfo{year}{2021}\natexlab{}.
\newblock \showarticletitle{Client {Selection} {Approach} in {Support} of
  {Clustered} {Federated} {Learning} over {Wireless} {Edge} {Networks}}. In
  \bibinfo{booktitle}{\emph{2021 {IEEE} {Global} {Communications} {Conference}
  ({GLOBECOM})}}. \bibinfo{pages}{1--6}.
\newblock
\urldef\tempurl%
\url{https://doi.org/10.1109/GLOBECOM46510.2021.9685938}
\showDOI{\tempurl}


\bibitem[Ali et~al\mbox{.}(2021)]%
        {ali_incentive-driven_2021}
\bibfield{author}{\bibinfo{person}{Asad Ali}, \bibinfo{person}{Inaam Ilahi},
  \bibinfo{person}{Adnan Qayyum}, \bibinfo{person}{Ihab Mohammed},
  \bibinfo{person}{Ala Al-Fuqaha}, {and} \bibinfo{person}{Junaid Qadir}.}
  \bibinfo{year}{2021}\natexlab{}.
\newblock \showarticletitle{Incentive-{Driven} {Federated} {Learning} and
  {Associated} {Security} {Challenges}: {A} {Systematic} {Review}}.
\newblock  (\bibinfo{year}{2021}), \bibinfo{pages}{30}.
\newblock


\bibitem[Anderson(2015)]%
        {anderson_technology_2015}
\bibfield{author}{\bibinfo{person}{Monica Anderson}.}
  \bibinfo{year}{2015}\natexlab{}.
\newblock \bibinfo{booktitle}{\emph{Technology device ownership: 2015}}.
\newblock \bibinfo{type}{Report}. \bibinfo{institution}{Pew Research Center}.
\newblock
\urldef\tempurl%
\url{https://apo.org.au/node/58353}
\showURL{%
\tempurl}


\bibitem[Antunes et~al\mbox{.}(2022)]%
        {antunes_federated_2022}
\bibfield{author}{\bibinfo{person}{Rodolfo~Stoffel Antunes},
  \bibinfo{person}{Cristiano André~da Costa}, \bibinfo{person}{Arne Küderle},
  \bibinfo{person}{Imrana~Abdullahi Yari}, {and} \bibinfo{person}{Björn
  Eskofier}.} \bibinfo{year}{2022}\natexlab{}.
\newblock \showarticletitle{Federated {Learning} for {Healthcare}: {Systematic}
  {Review} and {Architecture} {Proposal}}.
\newblock \bibinfo{journal}{\emph{ACM Transactions on Intelligent Systems and
  Technology}} \bibinfo{volume}{13}, \bibinfo{number}{4} (\bibinfo{date}{May}
  \bibinfo{year}{2022}), \bibinfo{pages}{54:1--54:23}.
\newblock
\showISSN{2157-6904}
\urldef\tempurl%
\url{https://doi.org/10.1145/3501813}
\showDOI{\tempurl}


\bibitem[Asad et~al\mbox{.}(2022)]%
        {asad_thf_2022}
\bibfield{author}{\bibinfo{person}{Muhammad Asad}, \bibinfo{person}{Ahmed
  Moustafa}, \bibinfo{person}{Fethi~A. Rabhi}, {and} \bibinfo{person}{Muhammad
  Aslam}.} \bibinfo{year}{2022}\natexlab{}.
\newblock \showarticletitle{{THF}: 3-{Way} {Hierarchical} {Framework} for
  {Efficient} {Client} {Selection} and {Resource} {Management} in {Federated}
  {Learning}}.
\newblock \bibinfo{journal}{\emph{IEEE Internet of Things Journal}}
  \bibinfo{volume}{9}, \bibinfo{number}{13} (\bibinfo{date}{July}
  \bibinfo{year}{2022}), \bibinfo{pages}{11085--11097}.
\newblock
\showISSN{2327-4662}
\urldef\tempurl%
\url{https://doi.org/10.1109/JIOT.2021.3126828}
\showDOI{\tempurl}
\newblock
\shownote{Conference Name: IEEE Internet of Things Journal}.


\bibitem[Balakrishnan et~al\mbox{.}(2022)]%
        {balakrishnan_diverse_2022}
\bibfield{author}{\bibinfo{person}{Ravikumar Balakrishnan},
  \bibinfo{person}{Tian Li}, \bibinfo{person}{Tianyi Zhou},
  \bibinfo{person}{Nageen Himayat}, \bibinfo{person}{Virginia Smith}, {and}
  \bibinfo{person}{Jeffrey Bilmes}.} \bibinfo{year}{2022}\natexlab{}.
\newblock \showarticletitle{{DIVERSE} {CLIENT} {SELECTION} {FOR} {FEDERATED}
  {LEARNING} {VIA} {SUBMODULAR} {MAXIMIZATION}}.
\newblock  (\bibinfo{year}{2022}), \bibinfo{pages}{18}.
\newblock


\bibitem[Cao et~al\mbox{.}(2022)]%
        {cao_c2s_2022}
\bibfield{author}{\bibinfo{person}{Mei Cao}, \bibinfo{person}{Yujie Zhang},
  \bibinfo{person}{Zezhong Ma}, {and} \bibinfo{person}{Mengying Zhao}.}
  \bibinfo{year}{2022}\natexlab{}.
\newblock \showarticletitle{{C2S}: {Class}-aware client selection for effective
  aggregation in federated learning}.
\newblock \bibinfo{journal}{\emph{High-Confidence Computing}}
  \bibinfo{volume}{2}, \bibinfo{number}{3} (\bibinfo{date}{Sept.}
  \bibinfo{year}{2022}), \bibinfo{pages}{100068}.
\newblock
\showISSN{2667-2952}
\urldef\tempurl%
\url{https://doi.org/10.1016/j.hcc.2022.100068}
\showDOI{\tempurl}


\bibitem[Cho et~al\mbox{.}(2022)]%
        {cho_towards_2022}
\bibfield{author}{\bibinfo{person}{Yae~Jee Cho}, \bibinfo{person}{Jianyu Wang},
  {and} \bibinfo{person}{Gauri Joshi}.} \bibinfo{year}{2022}\natexlab{}.
\newblock \showarticletitle{Towards {Understanding} {Biased} {Client}
  {Selection} in {Federated} {Learning}}. In
  \bibinfo{booktitle}{\emph{Proceedings of {The} 25th {International}
  {Conference} on {Artificial} {Intelligence} and {Statistics}}}.
  \bibinfo{publisher}{PMLR}, \bibinfo{pages}{10351--10375}.
\newblock
\urldef\tempurl%
\url{https://proceedings.mlr.press/v151/jee-cho22a.html}
\showURL{%
\tempurl}
\newblock
\shownote{ISSN: 2640-3498}.


\bibitem[Deng et~al\mbox{.}(2022)]%
        {deng_auction_2022}
\bibfield{author}{\bibinfo{person}{Yongheng Deng}, \bibinfo{person}{Feng Lyu},
  \bibinfo{person}{Ju Ren}, \bibinfo{person}{Huaqing Wu},
  \bibinfo{person}{Yuezhi Zhou}, \bibinfo{person}{Yaoxue Zhang}, {and}
  \bibinfo{person}{Xuemin Shen}.} \bibinfo{year}{2022}\natexlab{}.
\newblock \showarticletitle{{AUCTION}: {Automated} and {Quality}-{Aware}
  {Client} {Selection} {Framework} for {Efficient} {Federated} {Learning}}.
\newblock \bibinfo{journal}{\emph{IEEE Transactions on Parallel and Distributed
  Systems}} \bibinfo{volume}{33}, \bibinfo{number}{8} (\bibinfo{date}{Aug.}
  \bibinfo{year}{2022}), \bibinfo{pages}{1996--2009}.
\newblock
\showISSN{1558-2183}
\urldef\tempurl%
\url{https://doi.org/10.1109/TPDS.2021.3134647}
\showDOI{\tempurl}
\newblock
\shownote{Conference Name: IEEE Transactions on Parallel and Distributed
  Systems}.


\bibitem[Du et~al\mbox{.}(2022)]%
        {du_-device_2022}
\bibfield{author}{\bibinfo{person}{Zhaoyang Du}, \bibinfo{person}{Celimuge Wu},
  \bibinfo{person}{Tsutomu Yoshinage}, \bibinfo{person}{Lei Zhong}, {and}
  \bibinfo{person}{Yusheng Ji}.} \bibinfo{year}{2022}\natexlab{}.
\newblock \showarticletitle{On-device federated learning with fuzzy logic based
  client selection}. In \bibinfo{booktitle}{\emph{Proceedings of the
  {Conference} on {Research} in {Adaptive} and {Convergent} {Systems}}}
  \emph{(\bibinfo{series}{{RACS} '22})}. \bibinfo{publisher}{Association for
  Computing Machinery}, \bibinfo{address}{New York, NY, USA},
  \bibinfo{pages}{64--70}.
\newblock
\showISBNx{978-1-4503-9398-0}
\urldef\tempurl%
\url{https://doi.org/10.1145/3538641.3561490}
\showDOI{\tempurl}


\bibitem[Feldman(2015)]%
        {feldman_computational_2015}
\bibfield{author}{\bibinfo{person}{Michael Feldman}.}
  \bibinfo{year}{2015}\natexlab{}.
\newblock \emph{\bibinfo{title}{Computational {Fairness}: {Preventing}
  {Machine}-{Learned} {Discrimination}}}.
\newblock Thesis.
\newblock
\urldef\tempurl%
\url{https://scholarship.tricolib.brynmawr.edu/handle/10066/17628}
\showURL{%
\tempurl}
\newblock
\shownote{Accepted: 2016-01-19T17:37:36Z}.


\bibitem[Guo et~al\mbox{.}(2021)]%
        {guo_wcl_2021}
\bibfield{author}{\bibinfo{person}{Yingya Guo}, \bibinfo{person}{Kai Huang},
  {and} \bibinfo{person}{Jianshan Chen}.} \bibinfo{year}{2021}\natexlab{}.
\newblock \showarticletitle{{WCL}: {Client} {Selection} in {Federated}
  {Learning} with a {Combination} of {Model} {Weight} {Divergence} and {Client}
  {Training} {Loss} for {Internet} {Traffic} {Classification}}.
\newblock \bibinfo{journal}{\emph{Wireless Communications and Mobile
  Computing}}  \bibinfo{volume}{2021} (\bibinfo{date}{Dec.}
  \bibinfo{year}{2021}), \bibinfo{pages}{e3381998}.
\newblock
\showISSN{1530-8669}
\urldef\tempurl%
\url{https://doi.org/10.1155/2021/3381998}
\showDOI{\tempurl}
\newblock
\shownote{Publisher: Hindawi}.


\bibitem[Higgins et~al\mbox{.}(2019)]%
        {higgins_cochrane_2019}
\bibfield{author}{\bibinfo{person}{Julian P.~T. Higgins},
  \bibinfo{person}{James Thomas}, \bibinfo{person}{Jacqueline Chandler},
  \bibinfo{person}{Miranda Cumpston}, \bibinfo{person}{Tianjing Li},
  \bibinfo{person}{Matthew~J. Page}, {and} \bibinfo{person}{Vivian~A. Welch}.}
  \bibinfo{year}{2019}\natexlab{}.
\newblock \bibinfo{booktitle}{\emph{Cochrane {Handbook} for {Systematic}
  {Reviews} of {Interventions}}}.
\newblock \bibinfo{publisher}{John Wiley \& Sons}.
\newblock
\showISBNx{978-1-119-53661-1}
\newblock
\shownote{Google-Books-ID: cTqyDwAAQBAJ}.


\bibitem[Hosseinzadeh et~al\mbox{.}(2022a)]%
        {hosseinzadeh_federated_2022}
\bibfield{author}{\bibinfo{person}{Mehdi Hosseinzadeh}, \bibinfo{person}{Atefeh
  Hemmati}, {and} \bibinfo{person}{Amir~Masoud Rahmani}.}
  \bibinfo{year}{2022}\natexlab{a}.
\newblock \showarticletitle{Federated learning-based {IoT}: {A} systematic
  literature review}.
\newblock \bibinfo{journal}{\emph{International Journal of Communication
  Systems}} \bibinfo{volume}{35}, \bibinfo{number}{11} (\bibinfo{year}{2022}),
  \bibinfo{pages}{e5185}.
\newblock
\showISSN{1099-1131}
\urldef\tempurl%
\url{https://doi.org/10.1002/dac.5185}
\showDOI{\tempurl}
\newblock
\shownote{\_eprint: https://onlinelibrary.wiley.com/doi/pdf/10.1002/dac.5185}.


\bibitem[Hosseinzadeh et~al\mbox{.}(2022b)]%
        {hosseinzadeh_communication-loss_2022}
\bibfield{author}{\bibinfo{person}{Minoo Hosseinzadeh},
  \bibinfo{person}{Nathaniel Hudson}, \bibinfo{person}{Sam Heshmati}, {and}
  \bibinfo{person}{Hana Khamfroush}.} \bibinfo{year}{2022}\natexlab{b}.
\newblock \showarticletitle{Communication-{Loss} {Trade}-{Off} in {Federated}
  {Learning}: {A} {Distributed} {Client} {Selection} {Algorithm}}. In
  \bibinfo{booktitle}{\emph{2022 {IEEE} 19th {Annual} {Consumer}
  {Communications} \& {Networking} {Conference} ({CCNC})}}.
  \bibinfo{pages}{1--6}.
\newblock
\urldef\tempurl%
\url{https://doi.org/10.1109/CCNC49033.2022.9700601}
\showDOI{\tempurl}
\newblock
\shownote{ISSN: 2331-9860}.


\bibitem[Hou et~al\mbox{.}(2021)]%
        {hou_systematic_2021}
\bibfield{author}{\bibinfo{person}{Dongkun Hou}, \bibinfo{person}{Jie Zhang},
  \bibinfo{person}{Ka~Lok Man}, \bibinfo{person}{Jieming Ma}, {and}
  \bibinfo{person}{Zitian Peng}.} \bibinfo{year}{2021}\natexlab{}.
\newblock \showarticletitle{A {Systematic} {Literature} {Review} of
  {Blockchain}-based {Federated} {Learning}: {Architectures}, {Applications}
  and {Issues}}. In \bibinfo{booktitle}{\emph{2021 2nd {Information}
  {Communication} {Technologies} {Conference} ({ICTC})}}.
  \bibinfo{pages}{302--307}.
\newblock
\urldef\tempurl%
\url{https://doi.org/10.1109/ICTC51749.2021.9441499}
\showDOI{\tempurl}


\bibitem[Huang et~al\mbox{.}(2022)]%
        {huang_stochastic_2022}
\bibfield{author}{\bibinfo{person}{Tiansheng Huang}, \bibinfo{person}{Weiwei
  Lin}, \bibinfo{person}{Li Shen}, \bibinfo{person}{Keqin Li}, {and}
  \bibinfo{person}{Albert~Y. Zomaya}.} \bibinfo{year}{2022}\natexlab{}.
\newblock \showarticletitle{Stochastic {Client} {Selection} for {Federated}
  {Learning} {With} {Volatile} {Clients}}.
\newblock \bibinfo{journal}{\emph{IEEE Internet of Things Journal}}
  \bibinfo{volume}{9}, \bibinfo{number}{20} (\bibinfo{date}{Oct.}
  \bibinfo{year}{2022}), \bibinfo{pages}{20055--20070}.
\newblock
\showISSN{2327-4662}
\urldef\tempurl%
\url{https://doi.org/10.1109/JIOT.2022.3172113}
\showDOI{\tempurl}
\newblock
\shownote{Conference Name: IEEE Internet of Things Journal}.


\bibitem[Huang et~al\mbox{.}(2021)]%
        {huang_efficiency-boosting_2021}
\bibfield{author}{\bibinfo{person}{Tiansheng Huang}, \bibinfo{person}{Weiwei
  Lin}, \bibinfo{person}{Wentai Wu}, \bibinfo{person}{Ligang He},
  \bibinfo{person}{Keqin Li}, {and} \bibinfo{person}{Albert~Y. Zomaya}.}
  \bibinfo{year}{2021}\natexlab{}.
\newblock \showarticletitle{An {Efficiency}-{Boosting} {Client} {Selection}
  {Scheme} for {Federated} {Learning} {With} {Fairness} {Guarantee}}.
\newblock \bibinfo{journal}{\emph{IEEE Transactions on Parallel and Distributed
  Systems}} \bibinfo{volume}{32}, \bibinfo{number}{7} (\bibinfo{date}{July}
  \bibinfo{year}{2021}), \bibinfo{pages}{1552--1564}.
\newblock
\showISSN{1558-2183}
\urldef\tempurl%
\url{https://doi.org/10.1109/TPDS.2020.3040887}
\showDOI{\tempurl}
\newblock
\shownote{Conference Name: IEEE Transactions on Parallel and Distributed
  Systems}.


\bibitem[Jee~Cho et~al\mbox{.}(2020)]%
        {jee_cho_bandit-based_2020}
\bibfield{author}{\bibinfo{person}{Yae Jee~Cho}, \bibinfo{person}{Samarth
  Gupta}, \bibinfo{person}{Gauri Joshi}, {and} \bibinfo{person}{Osman Yağan}.}
  \bibinfo{year}{2020}\natexlab{}.
\newblock \showarticletitle{Bandit-based {Communication}-{Efficient} {Client}
  {Selection} {Strategies} for {Federated} {Learning}}. In
  \bibinfo{booktitle}{\emph{2020 54th {Asilomar} {Conference} on {Signals},
  {Systems}, and {Computers}}}. \bibinfo{pages}{1066--1069}.
\newblock
\urldef\tempurl%
\url{https://doi.org/10.1109/IEEECONF51394.2020.9443523}
\showDOI{\tempurl}
\newblock
\shownote{ISSN: 2576-2303}.


\bibitem[Kitchenham and Charters(2007)]%
        {kitchenham_guidelines_2007}
\bibfield{author}{\bibinfo{person}{Barbara Kitchenham} {and}
  \bibinfo{person}{Stuart Charters}.} \bibinfo{year}{2007}\natexlab{}.
\newblock \showarticletitle{Guidelines for performing {Systematic} {Literature}
  {Reviews} in {Software} {Engineering}}.
\newblock   \bibinfo{volume}{2} (\bibinfo{date}{Jan.} \bibinfo{year}{2007}).
\newblock


\bibitem[Ko et~al\mbox{.}(2021)]%
        {ko_joint_2021}
\bibfield{author}{\bibinfo{person}{Haneul Ko}, \bibinfo{person}{Jaewook Lee},
  \bibinfo{person}{Sangwon Seo}, \bibinfo{person}{Sangheon Pack}, {and}
  \bibinfo{person}{Victor C.~M. Leung}.} \bibinfo{year}{2021}\natexlab{}.
\newblock \showarticletitle{Joint {Client} {Selection} and {Bandwidth}
  {Allocation} {Algorithm} for {Federated} {Learning}}.
\newblock \bibinfo{journal}{\emph{IEEE Transactions on Mobile Computing}}
  (\bibinfo{year}{2021}), \bibinfo{pages}{1--1}.
\newblock
\showISSN{1558-0660}
\urldef\tempurl%
\url{https://doi.org/10.1109/TMC.2021.3136611}
\showDOI{\tempurl}
\newblock
\shownote{Conference Name: IEEE Transactions on Mobile Computing}.


\bibitem[Kuang et~al\mbox{.}(2021)]%
        {kuang_client_2021}
\bibfield{author}{\bibinfo{person}{Junqian Kuang}, \bibinfo{person}{Miao Yang},
  \bibinfo{person}{Hongbin Zhu}, {and} \bibinfo{person}{Hua Qian}.}
  \bibinfo{year}{2021}\natexlab{}.
\newblock \showarticletitle{Client {Selection} with {Bandwidth} {Allocation} in
  {Federated} {Learning}}. In \bibinfo{booktitle}{\emph{2021 {IEEE} {Global}
  {Communications} {Conference} ({GLOBECOM})}}. \bibinfo{pages}{01--06}.
\newblock
\urldef\tempurl%
\url{https://doi.org/10.1109/GLOBECOM46510.2021.9685090}
\showDOI{\tempurl}


\bibitem[Lee et~al\mbox{.}(2022)]%
        {lee_data_2022}
\bibfield{author}{\bibinfo{person}{Jaewook Lee}, \bibinfo{person}{Haneul Ko},
  \bibinfo{person}{Sangwon Seo}, {and} \bibinfo{person}{Sangheon Pack}.}
  \bibinfo{year}{2022}\natexlab{}.
\newblock \showarticletitle{Data {Distribution}-{Aware} {Online} {Client}
  {Selection} {Algorithm} for {Federated} {Learning} in {Heterogeneous}
  {Networks}}.
\newblock \bibinfo{journal}{\emph{IEEE Transactions on Vehicular Technology}}
  (\bibinfo{year}{2022}), \bibinfo{pages}{1--11}.
\newblock
\showISSN{1939-9359}
\urldef\tempurl%
\url{https://doi.org/10.1109/TVT.2022.3205307}
\showDOI{\tempurl}
\newblock
\shownote{Conference Name: IEEE Transactions on Vehicular Technology}.


\bibitem[Li et~al\mbox{.}(2022b)]%
        {li_pyramidfl_2022}
\bibfield{author}{\bibinfo{person}{Chenning Li}, \bibinfo{person}{Zeng Xiao},
  \bibinfo{person}{Mi Zhang}, {and} \bibinfo{person}{Zhichao Cao}.}
  \bibinfo{year}{2022}\natexlab{b}.
\newblock \bibinfo{title}{{PyramidFL} {\textbar} {Proceedings} of the 28th
  {Annual} {International} {Conference} on {Mobile} {Computing} {And}
  {Networking}}.
\newblock
\newblock
\urldef\tempurl%
\url{https://dl.acm.org/doi/10.1145/3495243.3517017}
\showURL{%
\tempurl}
\newblock
\shownote{Archive Location: world}.


\bibitem[Li et~al\mbox{.}(2022c)]%
        {li_uncertainty_2022}
\bibfield{author}{\bibinfo{person}{Pengfei Li}, \bibinfo{person}{Yunfeng Zhao},
  \bibinfo{person}{Liandong Chen}, \bibinfo{person}{Kai Cheng},
  \bibinfo{person}{Chuyue Xie}, \bibinfo{person}{Xiaofei Wang}, {and}
  \bibinfo{person}{Qinghua Hu}.} \bibinfo{year}{2022}\natexlab{c}.
\newblock \showarticletitle{Uncertainty {Measured} {Active} {Client}
  {Selection} for {Federated} {Learning} in {Smart} {Grid}}. In
  \bibinfo{booktitle}{\emph{2022 {IEEE} {International} {Conference} on {Smart}
  {Internet} of {Things} ({SmartIoT})}}. \bibinfo{pages}{148--153}.
\newblock
\urldef\tempurl%
\url{https://doi.org/10.1109/SmartIoT55134.2022.00032}
\showDOI{\tempurl}
\newblock
\shownote{ISSN: 2770-2677}.


\bibitem[Li et~al\mbox{.}(2020a)]%
        {li_federated_2020}
\bibfield{author}{\bibinfo{person}{Tian Li}, \bibinfo{person}{Anit~Kumar Sahu},
  \bibinfo{person}{Ameet Talwalkar}, {and} \bibinfo{person}{Virginia Smith}.}
  \bibinfo{year}{2020}\natexlab{a}.
\newblock \showarticletitle{Federated {Learning}: {Challenges}, {Methods}, and
  {Future} {Directions}}.
\newblock \bibinfo{journal}{\emph{IEEE Signal Processing Magazine}}
  \bibinfo{volume}{37}, \bibinfo{number}{3} (\bibinfo{date}{May}
  \bibinfo{year}{2020}), \bibinfo{pages}{50--60}.
\newblock
\showISSN{1558-0792}
\urldef\tempurl%
\url{https://doi.org/10.1109/MSP.2020.2975749}
\showDOI{\tempurl}
\newblock
\shownote{Conference Name: IEEE Signal Processing Magazine}.


\bibitem[Li et~al\mbox{.}(2020b)]%
        {li_fair_2020}
\bibfield{author}{\bibinfo{person}{Tian Li}, \bibinfo{person}{Maziar Sanjabi},
  \bibinfo{person}{Ahmad Beirami}, {and} \bibinfo{person}{Virginia Smith}.}
  \bibinfo{year}{2020}\natexlab{b}.
\newblock \bibinfo{title}{Fair {Resource} {Allocation} in {Federated}
  {Learning}}.
\newblock
\newblock
\urldef\tempurl%
\url{https://doi.org/10.48550/arXiv.1905.10497}
\showDOI{\tempurl}
\newblock
\shownote{arXiv:1905.10497 [cs, stat]}.


\bibitem[Li et~al\mbox{.}(2022a)]%
        {li_data_2022}
\bibfield{author}{\bibinfo{person}{Zonghang Li}, \bibinfo{person}{Yihong He},
  \bibinfo{person}{Hongfang Yu}, \bibinfo{person}{Jiawen Kang},
  \bibinfo{person}{Xiaoping Li}, \bibinfo{person}{Zenglin Xu}, {and}
  \bibinfo{person}{Dusit Niyato}.} \bibinfo{year}{2022}\natexlab{a}.
\newblock \showarticletitle{Data {Heterogeneity}-{Robust} {Federated}
  {Learning} via {Group} {Client} {Selection} in {Industrial} {IoT}}.
\newblock \bibinfo{journal}{\emph{IEEE Internet of Things Journal}}
  \bibinfo{volume}{9}, \bibinfo{number}{18} (\bibinfo{date}{Sept.}
  \bibinfo{year}{2022}), \bibinfo{pages}{17844--17857}.
\newblock
\showISSN{2327-4662}
\urldef\tempurl%
\url{https://doi.org/10.1109/JIOT.2022.3161943}
\showDOI{\tempurl}
\newblock
\shownote{Conference Name: IEEE Internet of Things Journal}.


\bibitem[Lin et~al\mbox{.}(2022)]%
        {lin_contribution-based_2022}
\bibfield{author}{\bibinfo{person}{Weiwei Lin}, \bibinfo{person}{Yinhai Xu},
  \bibinfo{person}{Bo Liu}, \bibinfo{person}{Dongdong Li},
  \bibinfo{person}{Tiansheng Huang}, {and} \bibinfo{person}{Fang Shi}.}
  \bibinfo{year}{2022}\natexlab{}.
\newblock \showarticletitle{Contribution-based {Federated} {Learning} client
  selection}.
\newblock \bibinfo{journal}{\emph{International Journal of Intelligent
  Systems}} \bibinfo{volume}{37}, \bibinfo{number}{10} (\bibinfo{year}{2022}),
  \bibinfo{pages}{7235--7260}.
\newblock
\showISSN{1098-111X}
\urldef\tempurl%
\url{https://doi.org/10.1002/int.22879}
\showDOI{\tempurl}
\newblock
\shownote{\_eprint: https://onlinelibrary.wiley.com/doi/pdf/10.1002/int.22879}.


\bibitem[Liu et~al\mbox{.}(2022)]%
        {liu_uplink_2022}
\bibfield{author}{\bibinfo{person}{Tingting Liu}, \bibinfo{person}{Haibo Zhou},
  \bibinfo{person}{Jun Li}, \bibinfo{person}{Feng Shu}, {and}
  \bibinfo{person}{Zhu Han}.} \bibinfo{year}{2022}\natexlab{}.
\newblock \showarticletitle{Uplink and {Downlink} {Decoupled} {5G}/{B5G}
  {Vehicular} {Networks}: {A} {Federated} {Learning} {Assisted} {Client}
  {Selection} {Method}}.
\newblock \bibinfo{journal}{\emph{IEEE Transactions on Vehicular Technology}}
  (\bibinfo{year}{2022}), \bibinfo{pages}{1--13}.
\newblock
\showISSN{1939-9359}
\urldef\tempurl%
\url{https://doi.org/10.1109/TVT.2022.3207916}
\showDOI{\tempurl}
\newblock
\shownote{Conference Name: IEEE Transactions on Vehicular Technology}.


\bibitem[Liu et~al\mbox{.}(2020)]%
        {liu_systematic_2020}
\bibfield{author}{\bibinfo{person}{Yi Liu}, \bibinfo{person}{Li Zhang},
  \bibinfo{person}{Ning Ge}, {and} \bibinfo{person}{Guanghao Li}.}
  \bibinfo{year}{2020}\natexlab{}.
\newblock \bibinfo{title}{A {Systematic} {Literature} {Review} on {Federated}
  {Learning}: {From} {A} {Model} {Quality} {Perspective}}.
\newblock
\newblock
\urldef\tempurl%
\url{https://doi.org/10.48550/arXiv.2012.01973}
\showDOI{\tempurl}
\newblock
\shownote{arXiv:2012.01973 [cs]}.


\bibitem[Lo et~al\mbox{.}(2021)]%
        {lo_systematic_2021}
\bibfield{author}{\bibinfo{person}{Sin~Kit Lo}, \bibinfo{person}{Qinghua Lu},
  \bibinfo{person}{Chen Wang}, \bibinfo{person}{Hye-Young Paik}, {and}
  \bibinfo{person}{Liming Zhu}.} \bibinfo{year}{2021}\natexlab{}.
\newblock \showarticletitle{A {Systematic} {Literature} {Review} on {Federated}
  {Machine} {Learning}: {From} a {Software} {Engineering} {Perspective}}.
\newblock \bibinfo{journal}{\emph{Comput. Surveys}} \bibinfo{volume}{54},
  \bibinfo{number}{5} (\bibinfo{date}{May} \bibinfo{year}{2021}),
  \bibinfo{pages}{95:1--95:39}.
\newblock
\showISSN{0360-0300}
\urldef\tempurl%
\url{https://doi.org/10.1145/3450288}
\showDOI{\tempurl}


\bibitem[Lo et~al\mbox{.}(2022)]%
        {lo_architectural_2022}
\bibfield{author}{\bibinfo{person}{Sin~Kit Lo}, \bibinfo{person}{Qinghua Lu},
  \bibinfo{person}{Liming Zhu}, \bibinfo{person}{Hye-Young Paik},
  \bibinfo{person}{Xiwei Xu}, {and} \bibinfo{person}{Chen Wang}.}
  \bibinfo{year}{2022}\natexlab{}.
\newblock \showarticletitle{Architectural patterns for the design of federated
  learning systems}.
\newblock \bibinfo{journal}{\emph{Journal of Systems and Software}}
  \bibinfo{volume}{191} (\bibinfo{date}{Sept.} \bibinfo{year}{2022}),
  \bibinfo{pages}{111357}.
\newblock
\showISSN{0164-1212}
\urldef\tempurl%
\url{https://doi.org/10.1016/j.jss.2022.111357}
\showDOI{\tempurl}


\bibitem[Ma et~al\mbox{.}(2021)]%
        {ma_client_2021}
\bibfield{author}{\bibinfo{person}{Jiahua Ma}, \bibinfo{person}{Xinghua Sun},
  \bibinfo{person}{Wenchao Xia}, \bibinfo{person}{Xijun Wang},
  \bibinfo{person}{Xiang Chen}, {and} \bibinfo{person}{Hongbo Zhu}.}
  \bibinfo{year}{2021}\natexlab{}.
\newblock \showarticletitle{Client {Selection} {Based} on {Label} {Quantity}
  {Information} for {Federated} {Learning}}. In \bibinfo{booktitle}{\emph{2021
  {IEEE} 32nd {Annual} {International} {Symposium} on {Personal}, {Indoor} and
  {Mobile} {Radio} {Communications} ({PIMRC})}}. \bibinfo{pages}{1--6}.
\newblock
\urldef\tempurl%
\url{https://doi.org/10.1109/PIMRC50174.2021.9569487}
\showDOI{\tempurl}
\newblock
\shownote{ISSN: 2166-9589}.


\bibitem[Ma et~al\mbox{.}(2022)]%
        {ma_state---art_2022}
\bibfield{author}{\bibinfo{person}{Xiaodong Ma}, \bibinfo{person}{Jia Zhu},
  \bibinfo{person}{Zhihao Lin}, \bibinfo{person}{Shanxuan Chen}, {and}
  \bibinfo{person}{Yangjie Qin}.} \bibinfo{year}{2022}\natexlab{}.
\newblock \showarticletitle{A state-of-the-art survey on solving non-{IID} data
  in {Federated} {Learning}}.
\newblock \bibinfo{journal}{\emph{Future Generation Computer Systems}}
  \bibinfo{volume}{135} (\bibinfo{date}{Oct.} \bibinfo{year}{2022}),
  \bibinfo{pages}{244--258}.
\newblock
\showISSN{0167-739X}
\urldef\tempurl%
\url{https://doi.org/10.1016/j.future.2022.05.003}
\showDOI{\tempurl}


\bibitem[McMahan et~al\mbox{.}(2017)]%
        {mcmahan_communication-efficient_2017}
\bibfield{author}{\bibinfo{person}{Brendan McMahan}, \bibinfo{person}{Eider
  Moore}, \bibinfo{person}{Daniel Ramage}, \bibinfo{person}{Seth Hampson},
  {and} \bibinfo{person}{Blaise Aguera~y Arcas}.}
  \bibinfo{year}{2017}\natexlab{}.
\newblock \showarticletitle{Communication-{Efficient} {Learning} of {Deep}
  {Networks} from {Decentralized} {Data}}. In
  \bibinfo{booktitle}{\emph{Proceedings of the 20th {International}
  {Conference} on {Artificial} {Intelligence} and {Statistics}}}.
  \bibinfo{publisher}{PMLR}, \bibinfo{pages}{1273--1282}.
\newblock
\urldef\tempurl%
\url{https://proceedings.mlr.press/v54/mcmahan17a.html}
\showURL{%
\tempurl}
\newblock
\shownote{ISSN: 2640-3498}.


\bibitem[Mohanani et~al\mbox{.}(2020)]%
        {mohanani_cognitive_2020}
\bibfield{author}{\bibinfo{person}{Rahul Mohanani}, \bibinfo{person}{Iflaah
  Salman}, \bibinfo{person}{Burak Turhan}, \bibinfo{person}{Pilar Rodríguez},
  {and} \bibinfo{person}{Paul Ralph}.} \bibinfo{year}{2020}\natexlab{}.
\newblock \showarticletitle{Cognitive {Biases} in {Software} {Engineering}: {A}
  {Systematic} {Mapping} {Study}}.
\newblock \bibinfo{journal}{\emph{IEEE Transactions on Software Engineering}}
  \bibinfo{volume}{46}, \bibinfo{number}{12} (\bibinfo{date}{Dec.}
  \bibinfo{year}{2020}), \bibinfo{pages}{1318--1339}.
\newblock
\showISSN{1939-3520}
\urldef\tempurl%
\url{https://doi.org/10.1109/TSE.2018.2877759}
\showDOI{\tempurl}
\newblock
\shownote{Conference Name: IEEE Transactions on Software Engineering}.


\bibitem[Nishio and Yonetani(2019)]%
        {nishio_client_2019}
\bibfield{author}{\bibinfo{person}{Takayuki Nishio} {and} \bibinfo{person}{Ryo
  Yonetani}.} \bibinfo{year}{2019}\natexlab{}.
\newblock \showarticletitle{Client {Selection} for {Federated} {Learning} with
  {Heterogeneous} {Resources} in {Mobile} {Edge}}. In
  \bibinfo{booktitle}{\emph{{ICC} 2019 - 2019 {IEEE} {International}
  {Conference} on {Communications} ({ICC})}}. \bibinfo{pages}{1--7}.
\newblock
\urldef\tempurl%
\url{https://doi.org/10.1109/ICC.2019.8761315}
\showDOI{\tempurl}
\newblock
\shownote{ISSN: 1938-1883}.


\bibitem[Pang et~al\mbox{.}(2022)]%
        {pang_incentive_2022}
\bibfield{author}{\bibinfo{person}{Jinlong Pang}, \bibinfo{person}{Jieling Yu},
  \bibinfo{person}{Ruiting Zhou}, {and} \bibinfo{person}{John~C.S. Lui}.}
  \bibinfo{year}{2022}\natexlab{}.
\newblock \showarticletitle{An {Incentive} {Auction} for {Heterogeneous}
  {Client} {Selection} in {Federated} {Learning}}.
\newblock \bibinfo{journal}{\emph{IEEE Transactions on Mobile Computing}}
  (\bibinfo{year}{2022}), \bibinfo{pages}{1--17}.
\newblock
\showISSN{1558-0660}
\urldef\tempurl%
\url{https://doi.org/10.1109/TMC.2022.3182876}
\showDOI{\tempurl}
\newblock
\shownote{Conference Name: IEEE Transactions on Mobile Computing}.


\bibitem[Pfitzner et~al\mbox{.}(2021)]%
        {pfitzner_federated_2021}
\bibfield{author}{\bibinfo{person}{Bjarne Pfitzner}, \bibinfo{person}{Nico
  Steckhan}, {and} \bibinfo{person}{Bert Arnrich}.}
  \bibinfo{year}{2021}\natexlab{}.
\newblock \showarticletitle{Federated {Learning} in a {Medical} {Context}: {A}
  {Systematic} {Literature} {Review}}.
\newblock \bibinfo{journal}{\emph{ACM Transactions on Internet Technology}}
  \bibinfo{volume}{21}, \bibinfo{number}{2} (\bibinfo{date}{June}
  \bibinfo{year}{2021}), \bibinfo{pages}{50:1--50:31}.
\newblock
\showISSN{1533-5399}
\urldef\tempurl%
\url{https://doi.org/10.1145/3412357}
\showDOI{\tempurl}


\bibitem[Qu et~al\mbox{.}(2022)]%
        {qu_context-aware_2022}
\bibfield{author}{\bibinfo{person}{Zhe Qu}, \bibinfo{person}{Rui Duan},
  \bibinfo{person}{Lixing Chen}, \bibinfo{person}{Jie Xu},
  \bibinfo{person}{Zhuo Lu}, {and} \bibinfo{person}{Yao Liu}.}
  \bibinfo{year}{2022}\natexlab{}.
\newblock \showarticletitle{Context-{Aware} {Online} {Client} {Selection} for
  {Hierarchical} {Federated} {Learning}}.
\newblock \bibinfo{journal}{\emph{IEEE Transactions on Parallel and Distributed
  Systems}} \bibinfo{volume}{33}, \bibinfo{number}{12} (\bibinfo{date}{Dec.}
  \bibinfo{year}{2022}), \bibinfo{pages}{4353--4367}.
\newblock
\showISSN{1558-2183}
\urldef\tempurl%
\url{https://doi.org/10.1109/TPDS.2022.3186960}
\showDOI{\tempurl}
\newblock
\shownote{Conference Name: IEEE Transactions on Parallel and Distributed
  Systems}.


\bibitem[Rai et~al\mbox{.}(2022)]%
        {rai_client_2022}
\bibfield{author}{\bibinfo{person}{Sumit Rai}, \bibinfo{person}{Arti Kumari},
  {and} \bibinfo{person}{Dilip~K. Prasad}.} \bibinfo{year}{2022}\natexlab{}.
\newblock \showarticletitle{Client {Selection} in {Federated} {Learning} under
  {Imperfections} in {Environment}}.
\newblock \bibinfo{journal}{\emph{AI}} \bibinfo{volume}{3}, \bibinfo{number}{1}
  (\bibinfo{date}{March} \bibinfo{year}{2022}), \bibinfo{pages}{124--145}.
\newblock
\showISSN{2673-2688}
\urldef\tempurl%
\url{https://doi.org/10.3390/ai3010008}
\showDOI{\tempurl}
\newblock
\shownote{Number: 1 Publisher: Multidisciplinary Digital Publishing Institute}.


\bibitem[Saha et~al\mbox{.}(2022)]%
        {saha_data-centric_2022}
\bibfield{author}{\bibinfo{person}{Rituparna Saha}, \bibinfo{person}{Sudip
  Misra}, \bibinfo{person}{Aishwariya Chakraborty},
  \bibinfo{person}{Chandranath Chatterjee}, {and} \bibinfo{person}{Pallav~Kumar
  Deb}.} \bibinfo{year}{2022}\natexlab{}.
\newblock \showarticletitle{Data-{Centric} {Client} {Selection} for {Federated}
  {Learning} over {Distributed} {Edge} {Networks}}.
\newblock \bibinfo{journal}{\emph{IEEE Transactions on Parallel and Distributed
  Systems}} (\bibinfo{year}{2022}), \bibinfo{pages}{1--12}.
\newblock
\showISSN{1558-2183}
\urldef\tempurl%
\url{https://doi.org/10.1109/TPDS.2022.3217271}
\showDOI{\tempurl}
\newblock
\shownote{Conference Name: IEEE Transactions on Parallel and Distributed
  Systems}.


\bibitem[Shaheen et~al\mbox{.}(2022)]%
        {shaheen_applications_2022}
\bibfield{author}{\bibinfo{person}{Momina Shaheen},
  \bibinfo{person}{Muhammad~Shoaib Farooq}, \bibinfo{person}{Tariq Umer}, {and}
  \bibinfo{person}{Byung-Seo Kim}.} \bibinfo{year}{2022}\natexlab{}.
\newblock \showarticletitle{Applications of {Federated} {Learning}; {Taxonomy},
  {Challenges}, and {Research} {Trends}}.
\newblock \bibinfo{journal}{\emph{Electronics}} \bibinfo{volume}{11},
  \bibinfo{number}{4} (\bibinfo{date}{Jan.} \bibinfo{year}{2022}),
  \bibinfo{pages}{670}.
\newblock
\showISSN{2079-9292}
\urldef\tempurl%
\url{https://doi.org/10.3390/electronics11040670}
\showDOI{\tempurl}
\newblock
\shownote{Number: 4 Publisher: Multidisciplinary Digital Publishing Institute}.


\bibitem[Shi et~al\mbox{.}(2022)]%
        {shi_vfedcs_2022}
\bibfield{author}{\bibinfo{person}{Fang Shi}, \bibinfo{person}{Chunchao Hu},
  \bibinfo{person}{Weiwei Lin}, \bibinfo{person}{Lisheng Fan},
  \bibinfo{person}{Tiansheng Huang}, {and} \bibinfo{person}{Wentai Wu}.}
  \bibinfo{year}{2022}\natexlab{}.
\newblock \showarticletitle{{VFedCS}: {Optimizing} {Client} {Selection} for
  {Volatile} {Federated} {Learning}}.
\newblock \bibinfo{journal}{\emph{IEEE Internet of Things Journal}}
  (\bibinfo{year}{2022}), \bibinfo{pages}{1--1}.
\newblock
\showISSN{2327-4662}
\urldef\tempurl%
\url{https://doi.org/10.1109/JIOT.2022.3195073}
\showDOI{\tempurl}
\newblock
\shownote{Conference Name: IEEE Internet of Things Journal}.


\bibitem[Tan et~al\mbox{.}(2022)]%
        {tan_reputation-aware_2022}
\bibfield{author}{\bibinfo{person}{Xavier Tan}, \bibinfo{person}{Wei~Chong Ng},
  \bibinfo{person}{Wei Yang~Bryan Lim}, \bibinfo{person}{Zehui Xiong},
  \bibinfo{person}{Dusit Niyato}, {and} \bibinfo{person}{Han Yu}.}
  \bibinfo{year}{2022}\natexlab{}.
\newblock \showarticletitle{Reputation-{Aware} {Federated} {Learning} {Client}
  {Selection} based on {Stochastic} {Integer} {Programming}}.
\newblock \bibinfo{journal}{\emph{IEEE Transactions on Big Data}}
  (\bibinfo{year}{2022}), \bibinfo{pages}{1--12}.
\newblock
\showISSN{2332-7790}
\urldef\tempurl%
\url{https://doi.org/10.1109/TBDATA.2022.3191332}
\showDOI{\tempurl}
\newblock
\shownote{Conference Name: IEEE Transactions on Big Data}.


\bibitem[van~den Berg et~al\mbox{.}(2013)]%
        {van_den_berg_overview_2013}
\bibfield{author}{\bibinfo{person}{Tobias van~den Berg},
  \bibinfo{person}{Martijn~W. Heymans}, \bibinfo{person}{Stephanie~S. Leone},
  \bibinfo{person}{David Vergouw}, \bibinfo{person}{Jill~A. Hayden},
  \bibinfo{person}{Arianne~P. Verhagen}, {and} \bibinfo{person}{Henrica~CW de
  Vet}.} \bibinfo{year}{2013}\natexlab{}.
\newblock \showarticletitle{Overview of data-synthesis in systematic reviews of
  studies on outcome prediction models}.
\newblock \bibinfo{journal}{\emph{BMC Medical Research Methodology}}
  \bibinfo{volume}{13}, \bibinfo{number}{1} (\bibinfo{date}{Dec.}
  \bibinfo{year}{2013}), \bibinfo{pages}{1--10}.
\newblock
\showISSN{1471-2288}
\urldef\tempurl%
\url{https://doi.org/10.1186/1471-2288-13-42}
\showDOI{\tempurl}
\newblock
\shownote{Number: 1 Publisher: BioMed Central}.


\bibitem[Wan et~al\mbox{.}(2022)]%
        {wan_shielding_2022}
\bibfield{author}{\bibinfo{person}{Wei Wan}, \bibinfo{person}{Shengshan Hu},
  \bibinfo{person}{Jianrong Lu}, \bibinfo{person}{Leo~Yu Zhang},
  \bibinfo{person}{Hai Jin}, {and} \bibinfo{person}{Yuanyuan He}.}
  \bibinfo{year}{2022}\natexlab{}.
\newblock \bibinfo{title}{Shielding {Federated} {Learning}: {Robust}
  {Aggregation} with {Adaptive} {Client} {Selection}}.
\newblock
\newblock
\urldef\tempurl%
\url{https://doi.org/10.48550/arXiv.2204.13256}
\showDOI{\tempurl}
\newblock
\shownote{arXiv:2204.13256 [cs]}.


\bibitem[Wang and Kantarci(2020)]%
        {wang_novel_2020}
\bibfield{author}{\bibinfo{person}{Yuwei Wang} {and} \bibinfo{person}{Burak
  Kantarci}.} \bibinfo{year}{2020}\natexlab{}.
\newblock \showarticletitle{A {Novel} {Reputation}-aware {Client} {Selection}
  {Scheme} for {Federated} {Learning} within {Mobile} {Environments}}. In
  \bibinfo{booktitle}{\emph{2020 {IEEE} 25th {International} {Workshop} on
  {Computer} {Aided} {Modeling} and {Design} of {Communication} {Links} and
  {Networks} ({CAMAD})}}. \bibinfo{pages}{1--6}.
\newblock
\urldef\tempurl%
\url{https://doi.org/10.1109/CAMAD50429.2020.9209263}
\showDOI{\tempurl}
\newblock
\shownote{ISSN: 2378-4873}.


\bibitem[Witt et~al\mbox{.}(2022)]%
        {witt_decentral_2022}
\bibfield{author}{\bibinfo{person}{Leon Witt}, \bibinfo{person}{Mathis Heyer},
  \bibinfo{person}{Kentaroh Toyoda}, \bibinfo{person}{Wojciech Samek}, {and}
  \bibinfo{person}{Dan Li}.} \bibinfo{year}{2022}\natexlab{}.
\newblock \bibinfo{title}{Decentral and {Incentivized} {Federated} {Learning}
  {Frameworks}: {A} {Systematic} {Literature} {Review}}.
\newblock
\newblock
\urldef\tempurl%
\url{https://doi.org/10.48550/arXiv.2205.07855}
\showDOI{\tempurl}
\newblock
\shownote{arXiv:2205.07855 [cs]}.


\bibitem[Wohlin(2014)]%
        {wohlin_guidelines_2014}
\bibfield{author}{\bibinfo{person}{Claes Wohlin}.}
  \bibinfo{year}{2014}\natexlab{}.
\newblock \showarticletitle{Guidelines for snowballing in systematic literature
  studies and a replication in software engineering}. In
  \bibinfo{booktitle}{\emph{Proceedings of the 18th {International}
  {Conference} on {Evaluation} and {Assessment} in {Software} {Engineering} -
  {EASE} '14}}. \bibinfo{publisher}{ACM Press}, \bibinfo{address}{London,
  England, United Kingdom}, \bibinfo{pages}{1--10}.
\newblock
\showISBNx{978-1-4503-2476-2}
\urldef\tempurl%
\url{https://doi.org/10.1145/2601248.2601268}
\showDOI{\tempurl}


\bibitem[Xu and Wang(2021)]%
        {xu_client_2021}
\bibfield{author}{\bibinfo{person}{Jie Xu} {and} \bibinfo{person}{Heqiang
  Wang}.} \bibinfo{year}{2021}\natexlab{}.
\newblock \showarticletitle{Client {Selection} and {Bandwidth} {Allocation} in
  {Wireless} {Federated} {Learning} {Networks}: {A} {Long}-{Term}
  {Perspective}}.
\newblock \bibinfo{journal}{\emph{IEEE Transactions on Wireless
  Communications}} \bibinfo{volume}{20}, \bibinfo{number}{2}
  (\bibinfo{date}{Feb.} \bibinfo{year}{2021}), \bibinfo{pages}{1188--1200}.
\newblock
\showISSN{1558-2248}
\urldef\tempurl%
\url{https://doi.org/10.1109/TWC.2020.3031503}
\showDOI{\tempurl}
\newblock
\shownote{Conference Name: IEEE Transactions on Wireless Communications}.


\bibitem[Yang et~al\mbox{.}(2022)]%
        {yang_client_2022}
\bibfield{author}{\bibinfo{person}{Miao Yang}, \bibinfo{person}{Hua Qian},
  \bibinfo{person}{Ximin Wang}, \bibinfo{person}{Yong Zhou}, {and}
  \bibinfo{person}{Hongbin Zhu}.} \bibinfo{year}{2022}\natexlab{}.
\newblock \showarticletitle{Client {Selection} for {Federated} {Learning}
  {With} {Label} {Noise}}.
\newblock \bibinfo{journal}{\emph{IEEE Transactions on Vehicular Technology}}
  \bibinfo{volume}{71}, \bibinfo{number}{2} (\bibinfo{date}{Feb.}
  \bibinfo{year}{2022}), \bibinfo{pages}{2193--2197}.
\newblock
\showISSN{1939-9359}
\urldef\tempurl%
\url{https://doi.org/10.1109/TVT.2021.3131852}
\showDOI{\tempurl}
\newblock
\shownote{Conference Name: IEEE Transactions on Vehicular Technology}.


\bibitem[Yu et~al\mbox{.}(2022)]%
        {yu_jointly_2022}
\bibfield{author}{\bibinfo{person}{Liangkun Yu}, \bibinfo{person}{Rana
  Albelaihi}, \bibinfo{person}{Xiang Sun}, \bibinfo{person}{Nirwan Ansari},
  {and} \bibinfo{person}{Michael Devetsikiotis}.}
  \bibinfo{year}{2022}\natexlab{}.
\newblock \showarticletitle{Jointly {Optimizing} {Client} {Selection} and
  {Resource} {Management} in {Wireless} {Federated} {Learning} for {Internet}
  of {Things}}.
\newblock \bibinfo{journal}{\emph{IEEE Internet of Things Journal}}
  \bibinfo{volume}{9}, \bibinfo{number}{6} (\bibinfo{date}{March}
  \bibinfo{year}{2022}), \bibinfo{pages}{4385--4395}.
\newblock
\showISSN{2327-4662}
\urldef\tempurl%
\url{https://doi.org/10.1109/JIOT.2021.3103715}
\showDOI{\tempurl}
\newblock
\shownote{Conference Name: IEEE Internet of Things Journal}.


\bibitem[Zeng et~al\mbox{.}(2020)]%
        {zeng_energy-efficient_2020}
\bibfield{author}{\bibinfo{person}{Qunsong Zeng}, \bibinfo{person}{Yuqing Du},
  \bibinfo{person}{Kaibin Huang}, {and} \bibinfo{person}{Kin~K. Leung}.}
  \bibinfo{year}{2020}\natexlab{}.
\newblock \showarticletitle{Energy-{Efficient} {Radio} {Resource} {Allocation}
  for {Federated} {Edge} {Learning}}. In \bibinfo{booktitle}{\emph{2020 {IEEE}
  {International} {Conference} on {Communications} {Workshops} ({ICC}
  {Workshops})}}. \bibinfo{pages}{1--6}.
\newblock
\urldef\tempurl%
\url{https://doi.org/10.1109/ICCWorkshops49005.2020.9145118}
\showDOI{\tempurl}
\newblock
\shownote{ISSN: 2474-9133}.


\bibitem[Zhang et~al\mbox{.}(2021c)]%
        {zhang_adaptive_2021}
\bibfield{author}{\bibinfo{person}{Hangjia Zhang}, \bibinfo{person}{Zhijun
  Xie}, \bibinfo{person}{Roozbeh Zarei}, \bibinfo{person}{Tao Wu}, {and}
  \bibinfo{person}{Kewei Chen}.} \bibinfo{year}{2021}\natexlab{c}.
\newblock \showarticletitle{Adaptive {Client} {Selection} in {Resource}
  {Constrained} {Federated} {Learning} {Systems}: {A} {Deep} {Reinforcement}
  {Learning} {Approach}}.
\newblock \bibinfo{journal}{\emph{IEEE Access}}  \bibinfo{volume}{9}
  (\bibinfo{year}{2021}), \bibinfo{pages}{98423--98432}.
\newblock
\showISSN{2169-3536}
\urldef\tempurl%
\url{https://doi.org/10.1109/ACCESS.2021.3095915}
\showDOI{\tempurl}
\newblock
\shownote{Conference Name: IEEE Access}.


\bibitem[Zhang et~al\mbox{.}(2021a)]%
        {zhang_dubhe_2021}
\bibfield{author}{\bibinfo{person}{Shulai Zhang}, \bibinfo{person}{Zirui Li},
  \bibinfo{person}{Quan Chen}, \bibinfo{person}{Wenli Zheng},
  \bibinfo{person}{Jingwen Leng}, {and} \bibinfo{person}{Minyi Guo}.}
  \bibinfo{year}{2021}\natexlab{a}.
\newblock \showarticletitle{Dubhe: {Towards} {Data} {Unbiasedness} with
  {Homomorphic} {Encryption} in {Federated} {Learning} {Client} {Selection}}.
  In \bibinfo{booktitle}{\emph{50th {International} {Conference} on {Parallel}
  {Processing}}}. \bibinfo{publisher}{ACM}, \bibinfo{address}{Lemont IL USA},
  \bibinfo{pages}{1--10}.
\newblock
\showISBNx{978-1-4503-9068-2}
\urldef\tempurl%
\url{https://doi.org/10.1145/3472456.3473513}
\showDOI{\tempurl}


\bibitem[Zhang et~al\mbox{.}(2021b)]%
        {zhang_client_2021}
\bibfield{author}{\bibinfo{person}{Wenyu Zhang}, \bibinfo{person}{Xiumin Wang},
  \bibinfo{person}{Pan Zhou}, \bibinfo{person}{Weiwei Wu}, {and}
  \bibinfo{person}{Xinglin Zhang}.} \bibinfo{year}{2021}\natexlab{b}.
\newblock \showarticletitle{Client {Selection} for {Federated} {Learning}
  {With} {Non}-{IID} {Data} in {Mobile} {Edge} {Computing}}.
\newblock \bibinfo{journal}{\emph{IEEE Access}}  \bibinfo{volume}{9}
  (\bibinfo{year}{2021}), \bibinfo{pages}{24462--24474}.
\newblock
\showISSN{2169-3536}
\urldef\tempurl%
\url{https://doi.org/10.1109/ACCESS.2021.3056919}
\showDOI{\tempurl}
\newblock
\shownote{Conference Name: IEEE Access}.


\bibitem[Zhu et~al\mbox{.}(2022)]%
        {zhu_client_2022}
\bibfield{author}{\bibinfo{person}{Hongbin Zhu}, \bibinfo{person}{Miao Yang},
  \bibinfo{person}{Junqian Kuang}, \bibinfo{person}{Hua Qian}, {and}
  \bibinfo{person}{Yong Zhou}.} \bibinfo{year}{2022}\natexlab{}.
\newblock \showarticletitle{Client {Selection} for {Asynchronous} {Federated}
  {Learning} with {Fairness} {Consideration}}. In
  \bibinfo{booktitle}{\emph{2022 {IEEE} {International} {Conference} on
  {Communications} {Workshops} ({ICC} {Workshops})}}.
  \bibinfo{pages}{800--805}.
\newblock
\urldef\tempurl%
\url{https://doi.org/10.1109/ICCWorkshops53468.2022.9814669}
\showDOI{\tempurl}
\newblock
\shownote{ISSN: 2694-2941}.


\end{thebibliography}

\end{document}